\documentclass[11pt]{article}

\usepackage{acl}

\usepackage{times}
\usepackage{latexsym}

\usepackage[T1]{fontenc}

\usepackage[utf8]{inputenc}

\usepackage{microtype}

\usepackage{inconsolata}

\usepackage{graphicx}

\usepackage{amsmath}
\usepackage{multirow}
\usepackage{booktabs}
\usepackage{enumitem}
\usepackage{algorithm}
\usepackage{algorithmic}

\usepackage{longtable}
\usepackage{booktabs}
\usepackage{array}
\usepackage{ragged2e}
\usepackage{xurl}
\usepackage{amssymb}

\newcommand{\VAR}[1]{{\small\url{#1}}}

\newcolumntype{C}[1]{>{\Centering\hspace{0pt}}p{#1}}
\newcolumntype{L}[1]{>{\RaggedRight\hspace{0pt}}p{#1}}

\title{Automatic Slide Updating with User-Defined Dynamic Templates and Natural Language Instructions}

\author{
  \textbf{Kun Zhou\textsuperscript{1,2}},
  \textbf{Jiakai He\textsuperscript{3}},
  \textbf{Wenmian Yang\textsuperscript{2,\textdagger}},
  \textbf{Zhensheng Wang\textsuperscript{1,2}},
  \textbf{Yiquan Zhang\textsuperscript{2}},
  \textbf{Weijia Jia\textsuperscript{2,4,\textdagger}}
  \\
  \textsuperscript{1}School of Artificial Intelligence, Beijing Normal University, Beijing, PR China\\
  \textsuperscript{2}Institute of Artificial Intelligence and Future Networks, Beijing Normal University, Zhuhai, PR China\\
  \textsuperscript{3}Faculty of Arts and Sciences, Beijing Normal University, Zhuhai, PR China\\
  \textsuperscript{4}Beijing Normal-Hong Kong Baptist University, Zhuhai, PR China\\
  \small{
    \{zhoukun, hejiakai, jensenwang\}@mail.bnu.edu.cn, 
    \{wenmianyang, jiawj\}@bnu.edu.cn, 
    zhangyq987@hotmail.com
  }
}

\begin{document}
\maketitle
\begingroup
\renewcommand\thefootnote{\textdagger}
\footnotetext{Corresponding authors.}
\endgroup
\begin{abstract}
Presentation slides are a primary medium for data-driven reporting, yet keeping complex, analytics-style decks up to date remains labor-intensive. Existing automation methods mostly follow fixed template filling and cannot support dynamic updates for diverse, user-authored slide decks. We therefore define ``Dynamic Slide Update via Natural Language Instructions on User-provided Templates'' and introduce DynaSlide, a large-scale benchmark with 20,036 real-world instruction–execution triples (source slide, user instruction, target slide) grounded in a shared external database and built from business reporting slides under bring-your-own-template (BYO-template) conditions. To tackle this task, we propose SlideAgent, an agent-based framework that combines multimodal slide parsing, natural language instruction grounding, and tool-augmented reasoning for tables, charts, and textual conclusions. SlideAgent updates content while preserving layout and style, providing a strong reference baseline on DynaSlide. We further design end-to-end and component-level evaluation protocols that reveal key challenges and opportunities for future research. The dataset and code are available at \url{https://github.com/XiaoZhou2024/SlideAgent}.
\end{abstract}

\section{Introduction}
Presentation slides are a core medium for integrating data, logical argumentation, and information dissemination in enterprise decision-making, market communication, and academic exchange \cite{fu2022doc2ppt,zheng2025pptagent,ge2025autopresent}. In data-driven settings such as business and operations analytics, high-quality slides must not only present key figures and conclusions accurately, but also preserve professional layout, narrative coherence, and stylistic consistency \cite{liao2025doclayllm,yang2025auto}. However, current slide production remains a labor-intensive, multi-stage process. From template selection, heterogeneous data collection, cleaning, and analysis, to chart design, insight writing, and final style unification, the workflow is cumbersome, requires specialized expertise, and is prone to error\cite{ge2025autopresent,chia2025m}.

A more critical bottleneck lies in the iterative updating of report-style slide decks. In periodic business reports (e.g., weekly, monthly, or quarterly), updates usually involve local data replacement and minor adjustments to conclusions, while the overall structure and visual style remain stable \cite{warner2023slidespecs,chen2025towards}. As a result, valuable human resources are consumed by low-value “copy–paste–modify” routines. This creates a clear motivation: reuse existing high-quality slide decks as personalized templates, and allow users to issue natural language instructions that drive automatic, dynamic data updates and content regeneration—while preserving the original design with high fidelity \cite{yang2025auto,ge2025autopresent, masry2025chartgemma}. Such a capability would substantially improve production efficiency and reduce human error\cite{long2024llms}.

Existing work falls short of this goal. Most related approaches adopt a fixed-template filling paradigm, where systems extract information from structured sources and inject it into rigid, pre-defined templates \cite{liao2025doclayllm}. These methods cannot handle diverse, user-authored slide decks with complex, idiosyncratic structures, and therefore fail to support realistic bring-your-own-template (BYO-template) scenarios. This gap limits the practical deployment of current document automation techniques \cite{livathinos2025docling,ouyang2025omnidocbench}.

In contrast to simple template filling, dynamic content update on user-provided slides is substantially more challenging \cite{faysse2024colpali,qin2026large}, which requires an end-to-end intelligent workflow. First, the system must perform strong multimodal document understanding, constructing a structured, element-level representation of arbitrary slide decks: identifying title hierarchies, body text, numerical values in tables, underlying data series in charts, and summary-level conclusions, together with their layout and dependencies\cite{verma2025graft,shen2022vila}. Second, the system must ground natural language update instructions into concrete, executable operations over this representation and the underlying data sources. Third and most challenging, it must achieve intelligent content synchronization between old and new data. This goes far beyond simple value substitution: it forms a closed-loop perception–reasoning–execution process that may involve external code execution to recompute statistics, charting engines to re-render visualizations, and language models to infer revised analytical conclusions\cite{masry2025chartgemma,lompo2025visual}. This tightly integrates multimodal understanding, user-intent modeling, data retrieval, tool invocation, and advanced reasoning into a technically demanding pipeline\cite{su2025difficulty,ding2024boosting}. To our knowledge, this direction has received limited attention in prior work.

To systematically address this gap, we formally define the task of Dynamic Slide Update via Natural Language Instructions on User-provided Templates. The task evaluates a system’s end-to-end ability, given an arbitrary user slide deck as a dynamic template, to (i) locate and query relevant data, (ii) update tables and charts, (iii) revise textual conclusions, and (iv) generate an updated slide deck that preserves the original design with high fidelity while reflecting the new data and instructions.

To support research on this task, we introduce DynaSlide, to our knowledge, the first benchmark for dynamic slide update under BYO-template conditions. Built on real-world real-estate analytics data from Beijing, Guangzhou, and Shenzhen (2020–2024), DynaSlide organizes raw data into structured PostgreSQL tables and curates diverse slide templates that mirror authentic business reporting scenarios. We construct 20,036 instruction–execution triples of the form ``source slide deck – user instruction – target slide deck,'' all grounded in a shared external PostgreSQL database that provides the underlying real-estate transaction records. Each triple is annotated with executable update instructions and corresponding data queries, covering a broad spectrum of business themes and statistical operations. We provide standardized train/validation/test splits and task variants to facilitate model development, ablation studies, and fine-grained evaluation.

To probe these challenges in a realistic setting and provide a strong reference point, we develop SlideAgent, an agent-based baseline system that automates slide updates via structured reasoning and tool orchestration. SlideAgent uses a large language model as a central controller that plans and coordinates specialized tools across two tightly coupled stages: (1) slide parsing and representation, where it constructs a hierarchical representation of slide elements and layout that captures semantic content, visual structure, and data dependencies; and (2) instruction-driven content synchronization, where it interprets natural language update instructions, maps them to data and document operations, calls external tools for data querying, computation, and chart rendering, and synthesizes revised textual content and conclusions. Throughout this process, SlideAgent maintains structured logs of its actions, which we use for reproducibility and fine-grained evaluation and error analysis on DynaSlide.

The main contributions of this paper are as follows:

\begin{itemize}
    \item We formally define Dynamic Slide Update via Natural Language Instructions on User-provided Templates and release DynaSlide, a large-scale benchmark with over 20k instruction–execution triples (``source slide – instruction – target slide'') grounded in a shared external database and constructed from realistic business reports.
    
    \item We develop SlideAgent, an agent-based system that orchestrates multimodal slide parsing, instruction grounding, and tool execution, providing a strong reference baseline for this task.

    \item We design end-to-end and module-level evaluation protocols on DynaSlide, and conduct detailed error analysis that reveals key challenges for future methods.
\end{itemize}

\section{Related Work}
\subsection{Automatic Slide Generation vs. Dynamic Update}

Research in presentation automation \cite{mondal2024presentations, bandyopadhyay2024enhancing} predominantly targets \textit{document-to-slide generation}, converting long-form inputs into slide decks. Recent systems such as SlideSpawn \cite{kumar2024slidespawn}, AutoPresent \cite{ge2025autopresent}, and PPTAgent \cite{zheng2025pptagent} extend this paradigm by incorporating layout planning and visual verification. However, these methods rely on fixed template libraries or generative layouts, treating slide creation as a one-off task. In contrast, our Dynamic Slide Update task addresses the novel challenge of updating user-provided templates while preserving their custom layouts and maintaining the logical dependencies between data, functions, and content elements.

\subsection{LLM Agents for Slide Automation}
LLM-based systems advance presentation automation through hierarchical text planning and multimodal parsing \cite{yue2024dots,yang2023mm,yao2022react}. 
Recent work integrates vision-language models for semantic understanding  \cite{pang2025paper2poster, ge2025autopresent} and modular tool invocation \cite{schick2023toolformer,qin2021co, zhang2023universal}. Yet these frameworks remain fundamentally limited: they update surface-level content (text, layout) but cannot reconstruct the underlying computational dependencies—table 
formulas, parameter propagation, and database queries—required for coherent multimodal updates. They lack support for dynamic, user-defined templates demanding coordinated multimodal updates. 
Our framework addresses this gap by integrating structured instruction parsing, multimodal layout analysis, and tool invocation to enable data-consistent updates across complex, real-world slides.

\section{DynaSlide Dataset}
DynaSlide is a large-scale benchmark designed to evaluate dynamic slide update on user-provided templates. Each instance is formulated as an instruction--execution task, consisting of (i) a source slide, (ii) a natural language update instruction, and (iii) a target slide generated by executing the instruction over a shared external PostgreSQL database of real-estate transactions. All slides are constructed from a family of controllable templates that factorize each slide into four element types, i.e., Title, Table, Chart, and Summary, and are grounded in executable metadata (SQL templates, statistical functions, and layout annotations) stored in YAML files. All textual content in DynaSlide (titles, captions, headers, and summaries) is in English; we design templates in the original language used in real-world reports and then translate them into English following a controlled protocol (see Appendix~\ref{sec:translation})


\subsection{Data Collection}
 \label{sec:Data Collection}
We collected large-scale residential transaction records spanning 2020--2024 from three major cities in China: Beijing, Guangzhou, and Shenzhen, covering both new and resale housing markets. Each record contains nine fields covering time, location, and transaction attributes (e.g., transaction date, city, block, project, supply volume, sales volume, transaction price, floor area, and unit price); detailed field descriptions are provided in Appendix~\ref{sec:Data Field Descriptions}. Following previous work \cite{wang2025retqa}, we source data from official and public real-estate transaction platforms, and apply three cleaning filters: (1) excluding records with missing critical attributes; (2) normalizing city and block names to ensure spatial consistency; and (3) retaining only blocks with at least 500 transactions to avoid sparsity. After cleaning, we obtain 1.79 million valid transaction records. These records are partitioned by city $\times$ market-type (new vs.\ resale), resulting in six relational tables. We release the processed records as a PostgreSQL database dump, which serves as the shared external data source for all slide generation and instruction execution.

\subsection{Template Design}
Since dynamic updates in real-world reporting primarily operate on titles, tables, charts, and summaries, we restrict DynaSlide to these four element types, which jointly cover textual, tabular, and analytical visual content. In consultation with real-estate domain experts, we determine six commonly used presentation themes and define 34 sub-templates as atomic layout units (see Appendix~\ref{sec:template specification}). Each sub-template specifies both the slide layout (element roles and positions) and the table-centric computation logic used to populate the slide.  The template and example of each element is illustrated in Figure~\ref{fig:template}, and the detailed descriptions of templates are shown in Appendix~\ref{sec:Template Design}.

\begin{figure}[t]
  \includegraphics[width=\linewidth]{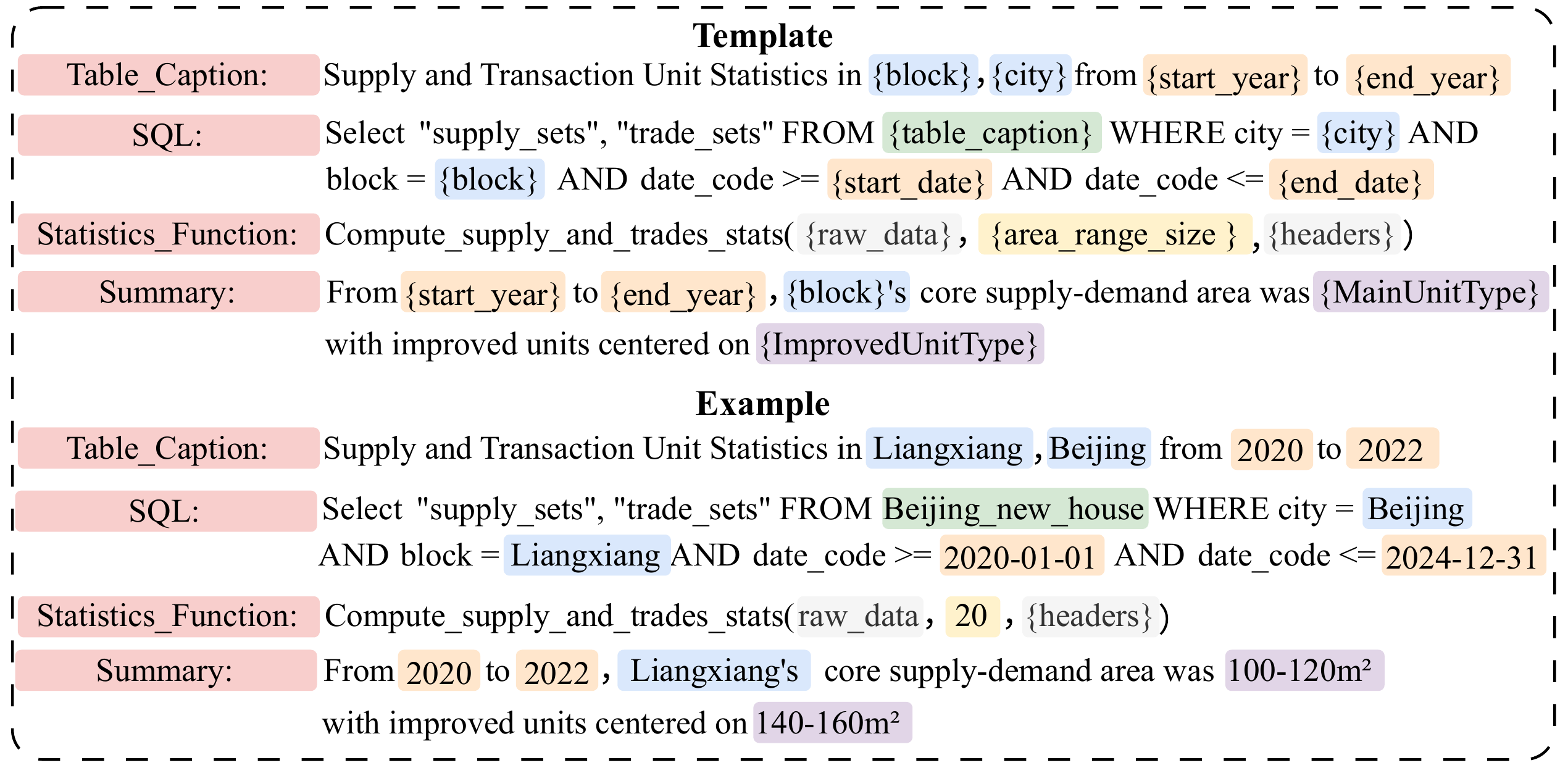}
  \caption{Illustration of the template filling.}
  \label{fig:template}
\end{figure}

\subsubsection{Title template design}
For each theme, we design three title templates that expose the statistical methodology and analytical perspective. During source slide generation, each instantiated sub-template randomly selects one title template to introduce lexical variation while preserving analytical intent.

\subsubsection{Table template design}
\label{sec:Table template design}
Tables consist of a caption and a table body. In practical business presentation, the table caption typically specifies the analysis topic and data scope, while the table body aggregates and statistically summarized results, e.g., shows monthly sales summaries across regions. For each theme, domain experts define a set of statistical analyses captured by 11 statistical functions $F_k$. Each table body is generated by pairing an SQL query template $SQL_k$ with a statistical function $F_k$: $SQL_k$ specifies which subset of records to retrieve from PostgreSQL under the slide’s scope, while $F_k$ specifies how to aggregate and summarize the retrieved data into table cells.

\paragraph{Caption generation.}
Captions describe the analytic scope (e.g., city/block and time span) using variable-based templates. As illustrated in Figure~\ref{fig:template}, each caption template combines fixed text (in black) with variable placeholders (in color). For each statistical function, we design three caption templates $C_p$ (33 in total).
We instantiate placeholders via a released ``Field Mapping Dictionary'' that maps template variables to concrete database fields and valid values (see Table~\ref{tab:field_mapping_dictionary} in Appendix~\ref{sec:Table_template_design}). Importantly, the same resolved variables are shared by the caption and the SQL template, ensuring that the caption’s semantics and the retrieved data are aligned.

\paragraph{Table body generation and linguistic diversity.}
After resolving variables, we populate $SQL_k$ to retrieve raw records, sample function-specific parameters from a released ``Parameter Candidate Dictionary'' (see Table~\ref{tab:parameter_dictionary} in Appendix~\ref{sec:Table_template_design}), and then apply $F_k$ to compute the table body. To diversify surface forms without changing semantics, we decouple header generation using a released ``{Header Alias Dictionary'' that maps canonical metrics to synonymous header variants (see Table~\ref{tab:header_alias_dictionary} in Appendix~\ref{sec:Table_template_design}). Computed outputs are normalized into canonical table structures for consistent rendering and evaluation (see Figure ~\ref{fig:table structure} in Appendix~\ref{sec:Table_template_design}).

\subsubsection{Chart template design}
Charts are visual projections of the same computed data used in tables (``homologous data, heterogeneous presentation''). We reuse $SQL_k$ and $F_k$ to obtain the underlying series and render bar/line charts according to the chart type specified by the sub-template.

\subsubsection{Summary template design}
Summaries synthesize computed tables into textual insights. For each $F_k$, we select and calculate 2-4 key metrics (e.g., trends, growth rates, peak values) from the table output. Based on these metrics, we select combinations of three sets of metrics, resulting in a total of 33 summary templates $Summary_q$. Similar to table generation, summary templates are instantiated by leveraging the Field Mapping Dictionary to resolve variable placeholders, ensuring semantic consistency across titles, captions, tables, charts, and summaries within a slide.

\subsubsection{Source slide Generation}
We implement ``EasySlide'', a toolkit built on \texttt{python-pptx}, to render titles, tables, charts, and summaries while enforcing consistent layout and typography. For each source slide, we release a YAML file recording (i) \texttt{slide\_filters} (database tables, $SQL_k$, $F_k$, resolved variables, and parameters) and (ii) \texttt{template\_slide} (element roles and layout), providing precise grounding for downstream dynamic updates. More details for ``EasySlide'' and source slide generation, see Appendix~\ref{sec:Source slide generation}. This systematic process yields 7,685 distinct source slides as the foundation for our dataset. By applying multiple update instructions to each source slide, we further construct the final instruction--execution benchmark.

\subsection{User Instruction and Target Slide Generation}
Each instruction--execution triple shares the same PostgreSQL database (Section~\ref{sec:Data Collection}). We design two update scenarios: (1) Basic Replacement, which updates only core scope variables (e.g., time span or region) while keeping $F_k$ unchanged; and (2) Customized-Parameter, which additionally modifies constraint parameters (e.g., area segmentation or price granularity), triggering cascading updates in both SQL retrieval and statistical computation. Instructions are sampled from 24 templates, and all variable changes are stored as \texttt{query\_filters} in YAML. Execution merges \texttt{slide\_filters} with \texttt{query\_filters} into \texttt{update\_filters}, re-queries PostgreSQL, recomputes the table (via $SQL_k$ and $F_k$), and synchronizes charts and summaries accordingly. Target slides are rendered to preserve the source slide’s layout and style. More details are shown in Appendix~\ref{sec:User Instruction and Target Slide Generation}.

\subsection{Dataset Statistics}
Our DynaSlide contains 20,036 instruction--execution samples. We split the dataset into train/validation/test with a 6:2:2 ratio (12{,}022 / 4{,}007 / 4{,}007), and enforce zero overlap of sub-templates across splits to evaluate generalization to unseen layouts. We release all source/target slides (PPTX), the PostgreSQL database dump, YAML metadata, and the three template dictionaries, enabling reproducible end-to-end and component-level evaluation. More details are shown in Appendix~\ref{sec:Dataset-Statistics}.

\section{Method}

\begin{figure*}[t]
\centering
  \includegraphics[width=\linewidth]{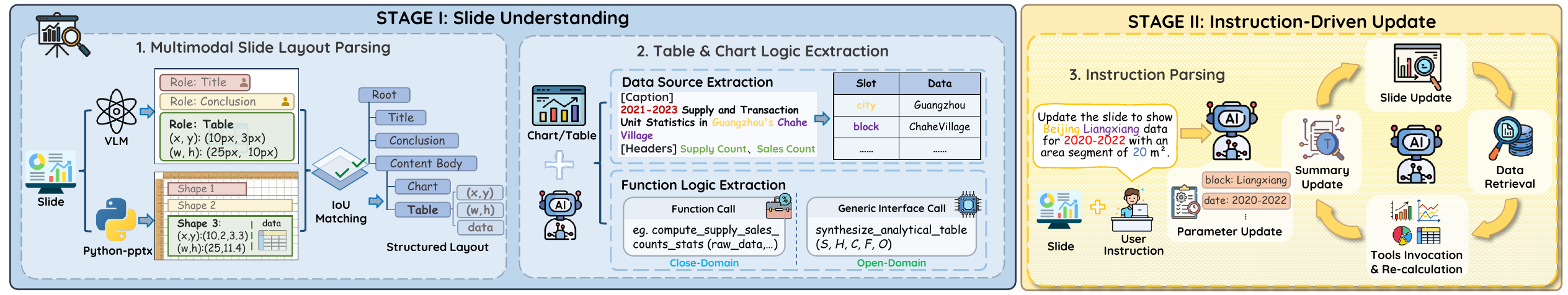}
  \caption{Overview of SlideAgent. The framework operates in two stages: (1) \textbf{Slide Understanding}, which parses layout and extracts logic; and (2) \textbf{Instruction-Driven Update}, which executes data queries and regenerates content.}
  \label{fig:Overview}
\end{figure*}

To address the challenges posed by DynaSlide, we develop SlideAgent, an agent-based baseline system for automated slide updates. As shown in Figure~\ref{fig:Overview}, SlideAgent adopts a two-stage architecture. Stage~1 (Slide Understanding) parses the input slide into a structured representation that captures element locations, data sources, and functional logic. Stage~2 (Instruction-Driven Update) interprets user instructions, retrieves updated data, executes the required transformations, and regenerates consistent textual, tabular, and visual content.


\subsection{Slide Parsing and Representation}
\label{sec:slide_parsing}
Accurate automated slide updates require both precise element localization and explicit recovery of the underlying data logic. We decompose this stage into two sub-tasks: multimodal layout parsing, which identifies semantic roles and spatial structures, and logic extraction, which recovers the data sources and aggregation logic for tables and charts.

\subsubsection{Multimodal Slide Layout Parsing}
\label{sec:layout_parsing}
Accurate identification of slide elements is essential for automated updates. However, programmatic slide libraries such as \texttt{python-pptx} provide limited semantic information, making it difficult to distinguish functional roles (e.g., titles, captions, or summaries) in complex layouts.

SlideAgent addresses this limitation by combining a vision--language model (VLM) with structured PPTX parsing provided by the \texttt{python-pptx} library to jointly recover semantic and geometric information. Specifically, each slide is first rendered as a PNG image and processed by Qwen-72B-VL using structured prompts (see Appendix~\ref{sec:appendix_prompts}). The VLM predicts semantic labels (e.g., title or table caption) along with approximate bounding boxes for visible elements. In parallel, \texttt{python-pptx} parses the underlying PPTX file to extract precise spatial coordinates and style metadata for all shape objects, as shown in Figure \ref{fig:Overview}.

We then align VLM predictions with PPTX shapes using an Intersection over Union (IoU) matching strategy~\cite{farhadi2018yolov3}. When the IoU exceeds a threshold of 0.5, the predicted semantic label is associated with the shape’s exact geometry and style attributes. The resulting structured representation records each element’s type, content, bounding box, and style information (Details see Appendix ~\ref{sec:Details of Structured Slide Representation}).


\subsubsection{Table and Chart Logic Extraction}
\label{sec:logic_extraction}
With accurately identified slide elements, in this section, we further reconstruct the underlying data logic for tables and charts, bridging the gap between the aggregated statistical data in the slides and the raw data. This requires reasoning over implicit data sources and statistical function logic. Since charts are visual representations of table data, we handle their logic extraction uniformly. This overall task presents two key challenges: (1) mapping aggregated statistics to the correct database sources and (2) recovering the aggregation logic used to generate them. We decompose this process into data source extraction and function logic extraction.


\paragraph{Data Source Extraction:}Recovering the raw data for tables and charts requires precise database queries. However, slides typically display high-level aggregated statistics (e.g., quarterly sales summaries), whereas databases store low-level granular records (e.g., individual transactions). Consequently, directly generating SQL from slide text is error-prone, since schema details of databases, such as table names and specific fields, are often implicit in the context of slides.

To address this, SlideAgent employs a schema-aligned slot-filling mechanism. We define nine standardized slots corresponding to the nine fields in the DynaSlide database (in Section~\ref{sec:Data Collection}). During extraction, schema cues are primarily drawn from two sources accessible via \texttt{python-pptx} in the slides: table captions, which encode analysis scope (e.g., time span or region), and table headers or chart series names, which indicate data dimensions. An LLM infers the target database table name and populates the corresponding slots based on the in-context learning (Details of prompts see Appendix~\ref{sec:appendix_prompts}), producing a structured JSON object that grounds each slide element to a concrete database context.
\label{sec:data_source_extraction}

\paragraph{Function Logic Extraction} To recover the computational logic transforming raw data into final table presentations, we distinguish two scenarios based on the availability of the underlying function definitions: a closed-domain scenario, and an open-domain scenario.

In the closed-domain scenario, we assume that tables are generated by a finite set of statistical functions known to the system. Specifically, SlideAgent maintains exact definitions for the 11 predefined functions described in Section~\ref{sec:Table template design}. These functions are exposed to the LLM as callable tools with explicit parameter schemas. Logic extraction is formulated as a function-calling task: given a target table, the LLM identifies the corresponding function and extracts its arguments. The recovered function and parameters are then used to directly invoke the intended computation logic. Details of tool descriptions and prompts see Appendix ~\ref{sec:appendix_prompts}.

\label{sec:function_logic_extraction}

In the open-domain scenario, the system has no prior knowledge of the specific function used to generate a table. This setting reflects real-world usage, where users may create arbitrary, custom analyses not covered by predefined templates. Although the specific generative function is unknown, the fundamental statistical methods (e.g., aggregation types, filtering metrics) remain analyzable. 

We therefore design a generic processing interface, \texttt{synthesize\_analytical\_table}, which reconstructs aggregation logic from atomic components parsed by the LLM. The interface operates on \textit{raw\_data} and is parameterized by five components:
\textit{(i) Table Structure Type}, categorizing the layout into canonical table structures (see Appendix~\ref{sec:Table_template_design});
\textit{(ii) Headers}, defining row and column dimensions;
\textit{(iii) Constraint Spec}, specifying binning or filtering logic with boundaries and step sizes;
\textit{(iv) Source Fields}, identifying required database columns (e.g., $\texttt{dim\_price}$);
and \textit{(v) Operations}, defining aggregation functions (e.g., SUM, AVG, COUNT).

The LLM decomposes the unknown logic into these parameters, which are assembled into an executable computation pipeline (see Algorithm~\ref{alg:generic_logic} in Appendix~\ref{sec:appendix_algorithm} for implementation details).

\subsection{Instruction-Driven Update}
\label{sec:instruction_update}
With the slide structure and logic extracted in Stage~1, Stage~2 performs dynamic updates following a four-step pipeline: instruction parsing, data retrieval, logic execution, and summary regeneration.

\subsubsection{User Instruction Parsing}
\label{sec:instruction_parsing}
User instructions specify updated data constraints or logic modifications in natural language. We model this step as a \textit{state update} task, as illustrated in Figure~\ref{fig:Overview}. The system takes the user instruction and the current parameter state (target table name, slots, and function parameters) as input, and prompts (see Appendix~\ref{sec:appendix_prompts}) an LLM to update relevant fields while strictly preserving the JSON schema. The output is a new parameter state that serves as the executable specification for downstream processing.

\subsubsection{SQL Generation and Data Retrieval}
\label{sec:sql_generation}
User instructions typically dictate specific data constraints or logic modifications. We thus model this task as a \textit{state update} problem, where the system takes the user instruction and the \textit{current parameter state} (a JSON object containing the target table name, slot values, and function logic parameters defined in Section~\ref{sec:logic_extraction}) as input. Using structured prompts (see Appendix~\ref{sec:appendix_prompts}), the LLM updates the relevant fields in the JSON object to reflect the user's instructions while preserving the schema structure. The output is a \textit{target parameter state} containing updated slots and function arguments, which serves as the executable specification for subsequent steps.

\subsubsection{Tool Invocation and Data Re-calculation}
\label{sec:tool_invocation}
Using the updated raw data and function parameters, SlideAgent re-executes the computation logic. Closed-domain tables invoke the corresponding predefined functions, while open-domain tables invoke \texttt{synthesize\_analytical\_table} with reconstructed logic parameters. Both cases yield updated aggregated statistical data.

\subsubsection{Summary Update and Final Rendering}
\label{sec:summary_update}
Since data updates may invalidate existing summaries, we introduce a fact-aware summary generation module. Given the original summary, the pre-update data, and the post-update data, the LLM rewrites the summary to reflect the updated trends. Details of the prompts are shown in Appendix ~\ref{sec:appendix_prompts}.

Finally, SlideAgent repopulates the slide with updated tables, charts, and summaries using the spatial coordinates and style metadata extracted in Stage~1 (Section~\ref{sec:slide_parsing}), ensuring strict preservation of the original layout and visual design.

\section{Experiments}
\label{sec:Experiments}
Dynamic slide update on user-provided templates is a newly introduced task and lacks directly comparable baselines. Accordingly, our experiments focus on: (1) comparing task success rates across different LLM backbones; (2) analyzing update accuracy across element types (Title, Table, Chart, Summary).

\subsection{Implementation Details}
We evaluate SlideAgent using five open-weight models as reasoning backbones: GPT-OSS (120B, 20B) and Qwen3-Instruct (80B, 30B, 14B). Qwen2.5-VL-72B is employed specifically for multimodal layout parsing. Detailed experimental configurations, including prompting strategies, software infrastructure (e.g., LangGraph, vLLM), and hardware specifications, are provided in Appendix~\ref{sec:implementation_details}.

\subsection{Evaluation Metrics}
We adopt two quantitative metrics. Task Success Rate (SR) is the primary metric, defined as the percentage of test cases in which the generated slide exactly matches the ground truth in both content (text and data values) and layout structure. To further analyze modality-specific performance, we report Element-Level Accuracy, measuring the update success rates for Titles, Tables, Charts, and Summaries individually.

\subsection{Main Results}

\begin{table*}[t]
\centering

\setlength{\tabcolsep}{3.5pt}
\small 
\begin{tabular}{lcccccccccccccc}
\toprule
\multirow{2}{*}{\textbf{Model}} & \multicolumn{2}{c}{\textbf{Theme 1}} & \multicolumn{2}{c}{\textbf{Theme 2}} & \multicolumn{2}{c}{\textbf{Theme 3}} & \multicolumn{2}{c}{\textbf{Theme 4}} & \multicolumn{2}{c}{\textbf{Theme 5}} & \multicolumn{2}{c}{\textbf{Theme 6}} & \multicolumn{2}{c}{\textbf{Average}} \\
\cmidrule(lr){2-3} \cmidrule(lr){4-5} \cmidrule(lr){6-7} \cmidrule(lr){8-9} \cmidrule(lr){10-11} \cmidrule(lr){12-13} \cmidrule(lr){14-15}
 & \textbf{Cls} & \textbf{Opn} & \textbf{Cls} & \textbf{Opn} & \textbf{Cls} & \textbf{Opn} & \textbf{Cls} & \textbf{Opn} & \textbf{Cls} & \textbf{Opn} & \textbf{Cls} & \textbf{Opn} & \textbf{Cls} & \textbf{Opn} \\
\midrule
GPT-OSS-120B & 90.12 & 78.29 & 71.83 & 70.91 & 78.48 & 62.56 & 77.03 & 58.17 & 81.01 & 72.31 & 85.36 & 70.89 & 80.64 & 68.86 \\
GPT-OSS-20B  & 81.48 & 68.45 & 61.55 & 57.14 & 65.95 & 49.62 & 61.69 & 44.17 & 67.09 & 60.40 & 77.45 & 57.72 & 69.20 & 56.25 \\
Qwen3-80B    & 87.65 & 75.69 & 70.14 & 65.68 & 73.42 & 60.27 & 67.53 & 53.67 & 69.62 & 62.84 & 83.64 & 65.32 & 75.33 & 63.91 \\
Qwen3-30B    & 85.06 & 71.06 & 63.38 & 61.50 & 68.35 & 54.95 & 63.64 & 48.00 & 68.35 & 61.26 & 79.64 & 61.39 & 71.40 & 59.69 \\
Qwen3-14B    & 50.74 & 29.09 & 50.70 & 29.09 & 34.18 & 25.11 & 38.96 & 31.67 & 45.57 & 35.01 & 52.73 & 37.22 & 45.48 & 31.13 \\
\bottomrule
\end{tabular}
\caption{Task Success Rates (\%) of SlideAgent across six presentation themes using different LLM backbones. Columns Cls and Opn denote Closed-Domain and Open-Domain update scenarios, respectively.}
\label{tab:overall-results}
\end{table*}
Table~\ref{tab:overall-results} reports the performance of SlideAgent across six presentation themes under both closed-domain and open-domain settings. The results yield three key observations regarding the complexity of dynamic slide updates.

First, model scale strongly correlates with task performance. GPT-OSS-120B achieves the highest success rates, reaching 80.64\% in closed-domain and 68.86\% in open-domain scenarios, exceeding its 20B counterpart by over 11\% and 12\%, respectively. A similar trend is observed for Qwen3 models, where Qwen3-80B (75.33\% / 63.91\%) substantially outperforms Qwen3-14B (45.48\% / 31.13\%). This large performance gap reflects the intrinsic difficulty of dynamic slide updates, which require a long chain of perception, querying, and computation. Errors in any intermediate step can invalidate the final output, and smaller models often fail to maintain strict instruction adherence over extended reasoning trajectories. In contrast, larger models demonstrate greater robustness in handling long-context, multi-step workflows.

Second, open-domain scenarios consistently degrade performance, disproportionately affecting smaller models. While GPT-OSS-120B shows a moderate 14.6\% relative decline (80.64\%$\to$68.86\%), Qwen3-14B exhibits a much larger 31.5\% relative decline (45.48\%$\to$31.13\%). This pattern aligns with the differing requirements of the two logic extraction modes: closed-domain update mainly involves selecting from a predefined function library, whereas open-domain update requires reconstructing computation logic from scratch, where larger models show stronger generalization beyond template memorization.

Third, task difficulty varies significantly across presentation themes. Taking GPT-OSS-120B as an example, Theme~1 achieves 90.12\% in closed-domain settings, whereas Themes~2, 3, and~4 pose substantially greater challenges, reaching only 71.83\%, 78.48\%, and 77.03\%, respectively.  This variation reflects the differing demands placed on the perception--querying--computation pipeline. Theme~1 primarily involves simple table structures and basic statistical operations, facilitating reliable layout parsing and field mapping. In contrast, Themes~2-~4 require complex cross-dimensional aggregations and multi-layer constraints, which amplify the likelihood of errors during logic extraction and SQL generation. These results indicate that, even under identical database schemas and model architectures, intrinsic task complexity remains a critical bottleneck for automation.

\begin{figure}[t]
  \includegraphics[width=\linewidth]{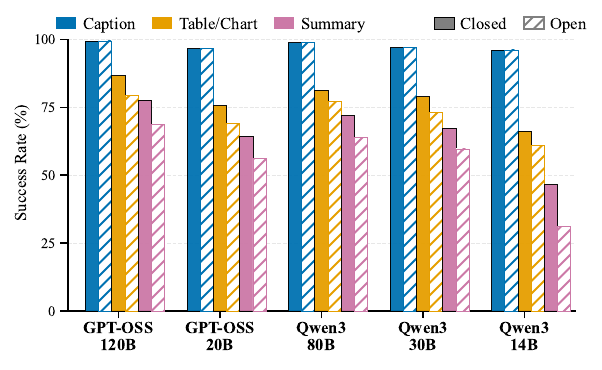}
  \caption{Element-level accuracy across closed-domain (Cls) and open-domain (Opn) scenarios.}
  \label{fig:res}
\end{figure}

Figure~\ref{fig:res} further breaks down performance by element type. Caption updates consistently achieve high accuracy across all models, as they largely involve direct substitution of constraint parameters (e.g., dates or locations) with minimal reasoning. In contrast, Table and Chart updates are substantially more challenging, with accuracy ranging from 66.23\% to 86.67\% in closed-domain settings and dropping to 60.82\%--79.24\% in open-domain settings. This degradation empirically supports the analysis in Section~\ref{sec:logic_extraction}: open-domain scenarios require models to reverse-engineer aggregation logic from visual and textual cues rather than selecting predefined functions, increasing error accumulation during SQL generation. Summary updates exhibit the highest failure rates, particularly for smaller models (e.g., Qwen3-14B drops to 37.61\% in open-domain settings), confirming that summary generation is a fact-aware reasoning task that requires strict consistency between recomputed values and the textual narrative—a capability that scales strongly with model size.

\begin{table*}[t]
\centering
\small 
\setlength{\tabcolsep}{5pt}
\begin{tabular}{lccccccc}
\toprule
\textbf{Model} & \textbf{Func. Logic} & \textbf{Data Src.} & \textbf{Instr. Parse} & \textbf{SQL Gen.} & \textbf{Tool Inv.} & \textbf{Sum. Upd.} & \textbf{Task SR$^\dagger$} \\ 
\midrule
GPT-OSS-120B & 88.34 & 90.37 & 85.69 & 85.56 & 79.24 & 68.44 & 68.86 \\
GPT-OSS-20B  & 74.63 & 77.96 & 76.44 & 76.21 & 69.13 & 56.30 & 56.25 \\
Qwen3-80B    & 87.19 & 86.14 & 82.63 & 82.43 & 77.16 & 64.72 & 63.91 \\
Qwen3-30B    & 81.55 & 82.40 & 78.59 & 79.54 & 73.20 & 60.51 & 59.69 \\
Qwen3-14B    & 63.38 & 70.17 & 70.35 & 73.09 & 60.82 & 31.25 & 31.13 \\
\bottomrule
\end{tabular}
\caption{Module-wise accuracy (\%) in open-domain scenarios. \textbf{Task SR$^\dagger$} denotes the end-to-end Task Success Rate from Table~\ref{tab:overall-results} for comparison. Abbreviations: \textbf{Func. Logic}: Function Logic Extraction, \textbf{Data Src.}: Data Source Extraction, \textbf{Instr. Parse}: User Instruction Parsing, \textbf{SQL Gen.}: SQL Generation, \textbf{Tool Inv.}: Tool Invocation, \textbf{Sum. Upd.}: Summary Update.}
\label{tab:open-domain-modules}
\end{table*}

To provide deeper insights into the performance bottlenecks of dynamic slide updates, we conduct module-wise evaluations that isolate the accuracy of individual pipeline components. This analysis enables fine-grained understanding of where current systems succeed and where they encounter fundamental challenges.

The Multimodal Slide Layout Parsing module achieves 99.5\% accuracy in our evaluation. Since this module consistently uses Qwen2.5-VL-72B for layout parsing across all settings, the layout parsing results remain identical for all model variants. Given that each PowerPoint slide typically contains only 4--6 elements, the layout structure is relatively simple, which also makes the parsing task more stable.

The accuracy of six core modules in open-domain scenarios is reported in Table~\ref{tab:open-domain-modules}. In this setting, the system is required to reconstruct computation logic without access to predefined function libraries. The evaluation assesses the fidelity of intermediate outputs at each stage of the SlideAgent pipeline, measuring how accurately each component executes its specific sub-task.

Several critical patterns emerge from these results. Function Logic Extraction and Data Source Extraction consistently achieve the highest accuracy across all models, with GPT-OSS-120B reaching 88.34\% and 90.37\% respectively. This indicates that larger models possess strong capabilities in reverse-engineering statistical operations and mapping slide content to database schemas.

In contrast, Summary Update presents the most significant challenge, with accuracy dropping substantially across all model scales. GPT-OSS-120B achieves only 68.44\% on this module, while smaller models like Qwen3-14B fall to 31.25\%. This confirms our hypothesis that summary generation requires sophisticated fact-aware reasoning to ensure textual narratives faithfully reflect updated data trends—a capability that scales strongly with model capacity.

The performance gap between Function Logic Extraction and Summary Update (19.90 percentage points for GPT-OSS-120B) reveals a key insight: while models can effectively extract and execute computational logic, translating quantitative updates into coherent natural language conclusions remains a fundamental bottleneck. This suggests that future work should prioritize developing specialized modules for cross-modal consistency checking and data-grounded text generation.

Model scale demonstrates consistent impact across all modules. Comparing GPT-OSS-120B to GPT-OSS-20B, we observe improvements of 13.71 points in Function Logic Extraction, 12.41 points in Data Source Extraction, and 12.14 points in Summary Update. These gains highlight that high-order reasoning tasks in dynamic slide updates benefit substantially from increased model capacity.

To further assess generalization beyond the main benchmark setting, we conduct two supplementary experiments and summarize a representative case study before concluding. First, on 500 randomly sampled template-generated cases, stronger models continue to outperform weaker ones across both closed and open settings, with closed-source models remaining consistently ahead of open-weight alternatives. The relative ranking among open-weight models is also preserved, indicating that the end-to-end slide updating task is strongly capability-dependent. Second, we compare the same sampled cases with human-authored slides and observe broadly consistent trends, suggesting that DynaSlide evaluates generalizable slide-updating ability rather than template-specific matching. We additionally analyze representative failure modes of SlideAgent through a case study. The results show that failures mainly stem from error accumulation across the perception--reasoning--execution pipeline and from the difficulty of maintaining numerically precise summary rewriting in the final stage. Together, these observations further highlight that the central challenge lies not only in executing each module correctly, but also in preserving cross-module consistency through to the final narrative. Detailed results and discussion are provided in Appendix~\ref{sec:appendix_generalization}, and Appendix~\ref{sec:appendix_failure_cases}.

\section{Conclusion}
In this paper, we formalize the task of ``Dynamic Slide Update via Natural Language Instructions on User-provided Templates'', addressing realistic bring-your-own-template scenarios that are not supported by existing template-based methods. To facilitate systematic research, we introduce DynaSlide, a large-scale benchmark with 20,036 real-world instruction–execution triples grounded in a shared external database. We further propose SlideAgent, an agent-based framework that combines multimodal slide parsing, instruction grounding, and tool-augmented reasoning to update tables, charts, and textual conclusions while preserving layout and style. Extensive evaluations reveal key challenges in logic reconstruction, cross-modal consistency, and fact-aware summary generation.

\section*{Limitations}
 This study has several limitations that do not undermine its contributions. First, our benchmark focuses exclusively on the real estate domain (e.g., housing transaction data, price trends, supply-demand analysis). This is not a limitation of capability, but a deliberate design to ensure data complexity: real estate data exhibits rich interdependencies—such as how price changes affect sales volume, or how regional supply impacts unit pricing—that are ideal for stress-testing an agent's numerical reasoning and multi-variable coordination. The core mechanism—interpreting natural language instructions, querying a database, updating tables/charts, and regenerating summaries—is domain-agnostic and directly applicable to other data-intensive sectors (e.g., financial earnings reports, supply chain dashboards, clinical trial summaries) where similar logic consistency is required. We chose depth over breadth: mastering one complex domain provides stronger evidence of reasoning capability than superficial coverage of many simple domains.

Second, DynaSlide employs a family of controllable templates rather than uncurated slides scraped from the web. This decision sacrifices some stylistic diversity to ensure executable ground truth. In a benchmark setting, verifiable correctness is essential; using arbitrary wild files would preclude the automated, deterministic evaluation of table values and chart series. Our templates thus serve as a standardized testbed where layout fidelity and content accuracy can be rigorously measured, avoiding the evaluation noise inherent in open-ended generation tasks. Future work may explore adapting our logic-driven update engine to noisier, unstructured layout parsers once the core reasoning framework is established.

Third, our framework operates on the premise that slide elements are grounded to a structured database (PostgreSQL). This alignment reflects standard enterprise workflows, where periodic reports (e.g., monthly reviews) are typically derived from pre-established data warehouses. Consequently, our system addresses the update phase of the reporting lifecycle. The "cold start" problem—reconstructing structured databases from legacy static presentations—constitutes a distinct document understanding challenge requiring chart de-rendering and schema inference, which are orthogonal to our focus on logically consistent updates. Our approach targets the high-frequency scenario of maintaining reports with accessible data lineages, a scope covering a significant portion of automated reporting needs.

Finally, our work prioritizes the analytical core of presentations: tables, charts, and data-driven summaries. We do not currently handle decorative graphics or conceptual diagrams. This scope is motivated by the observation that in periodic reporting, analytical elements account for the vast majority (>90\%) of content modifications. By focusing on these high-frequency update targets, we achieve broad multimodal coverage—spanning numerical reasoning, visual rendering, and natural language generation—while maintaining engineering tractability. Supporting non-analytical visual content would require additional perception modules (e.g., for semantic style transfer) that are complementary to, but distinct from, the logical update problem addressed here.

\section*{Acknowledgments}
This work is supported in part by the National Natural Science Foundation
of China (NSFC) under Grant 62272050 and the grant of Beijing Normal-
Hong Kong Baptist University sponsored
by Guangdong Provincial Department of Education; in part by Zhuhai Science-Tech Innovation Bureau under
Grant No. 2320004002772 and the Interdisciplinary Intelligence
Super Computer Center of Beijing Normal University (Zhuhai).

\bibliography{custom}

\appendix
\section{Dataset Construction}
\subsection{Translation Protocol}
\label{sec:translation}
Since the original slide templates and data schema were derived from the Chinese real estate market, all textual elements were initially in Chinese. We employed a three-step translation protocol to convert them into English:

\textbf{Template Content Translation.} We translated all static text in titles, captions, and summaries from Chinese to English, while preserving dynamic placeholder variables.

\textbf{Database Schema and Dictionary Adaptation.} We translated database schema names (table and column names) into English. For the Header Alias Dictionary, we curated diverse English synonyms for each metric (e.g., 'Avg Price', 'Unit Cost', 'Average Value' for the same price field), enabling linguistic variety in the final slides. The Field Mapping Dictionary was updated to link English template variables to database fields.

\textbf{Terminology Verification.} Domain experts reviewed all translated assets to ensure technical terms matched professional English standards in real estate.

\subsection{Data Field Descriptions}
\label{sec:Data Field Descriptions}
 The detailed schema of the real-estate database used in this study is presented in Table \ref{tab:field_descriptions_database}. This dataset compiles residential transaction records from major Chinese cities across both new and resale housing markets. Each entry contains nine key attributes that jointly encode spatial location (e.g., city, block), temporal information, and core transaction metrics (e.g., volume, floor area, price). These fields constitute the essential data backbone that underpins the statistical analyses and automated reporting workflows presented in the main text.

\begin{table*}[h]
  \centering
  \small
  \setlength{\tabcolsep}{4pt} 
  \renewcommand{\arraystretch}{1.2}
  \caption{Description of fields in the real-estate database.}
  \label{tab:field_descriptions_database}
  \begin{tabular}{p{2.2cm} p{1.8cm} p{7.0cm} p{3.3cm}}
    \toprule
    \textbf{Field Name} & \textbf{Data Type} & \textbf{Definition} & \textbf{Example} \\
    \midrule
    city & String & Name of the target city. & Beijing, Shanghai \\
    block & String & Name of the administrative district or sub-region. & Chaoyang District \\
    project & String & Name of the residential project or community. & Oceanwide Elite Estate \\
    date\_code & String & Temporal identifier. & 2024-05-01 \\
    supply\_sets & Integer & Volume of new housing units listed for sale. & 1 \\
    trade\_sets & Integer & Volume of housing units sold/transacted. & 1 \\
    dim\_area & Float & Floor area (\(m^2\)). & 125.5 \\
    dim\_price & Float & Total price (15.2k CNY). & 15.2 \\
    dim\_unit\_price & Float & Price per square meter (CNY/\(m^2\)). & 68000.0 \\
    \bottomrule
  \end{tabular}
\end{table*}

\subsection{Template Specification}
\label{sec:template specification}

To ensure reproducibility and facilitate fine-grained evaluation, we provide the full specification of our template system in Table~\ref{tab:full_templates}. This table enumerates the 34 atomic sub-templates across all identified business themes. Each entry explicitly defines the four core elements required for slide generation: (1) a set of \textbf{Title Templates} that introduce the analytical perspective; (2) a specific \textbf{Statistical Function} ($F_k$) that drives data retrieval and aggregation; (3) \textbf{Caption Templates} that define the scope of tables/charts; and (4) \textbf{Summary Templates} that synthesize quantitative insights into natural language. All templates utilize variable placeholders (e.g., \texttt{\VAR{City}}, \texttt{\VAR{Start\_Year}}) denoted in \texttt{monospace}. These variables are automatically populated during generation using the real-estate market variable definitions provided in Table~\ref{tab:realestatevariables}, ensuring semantic consistency across all slide elements.

\subsection{Template Design}
\label{sec:Template Design}
In professional real-estate presentations, slides typically consist of several key elements: titles, body text, tables, charts, images, and summary statements. For the slide auto-update task, we focus specifically on four core element types—Title, Table, Chart, and Summary—for two reasons: (1) they are the most frequently updated in real-world reporting workflows; (2) together they cover all principal content modalities (text, tabular data, visual analytics).
Building on thorough domain analysis, we identify six recurring business analysis themes (see Appendix~\ref{sec:template specification}), such as \emph{Cross-Structure Analysis of Resale-House Transactions} and \emph{Analysis of Neighborhood Price Trends}. For each theme $M_i$, we define a prototype slide composed of 2–7 configurable sub-templates $T_j$, yielding a total of 34 distinct sub-templates tailored to common real-estate analytic scenarios. These 34 sub-templates serve as the atomic layout units that we later use to build source slides and to define train/validation/test splits.
The design logic for Title, Table, Chart, and Summary templates is detailed below.
\subsubsection{Title template design}
Titles in presentations serve as both guides and indices, directly exposing the statistical methodology and analytical perspective employed on each slide. For example, the title ``Cross-Analysis of New Housing Transaction Structure'' signals that the slide presents multi-dimensional cross-tabulated statistics.
In designing the $\textit{Title}_i$ template suite, we created three title templates for each analytical theme (18 in total; see Appendix~\ref{sec:template specification}). During source slide generation, for each sub-template $T_j$ within theme $M_i$, the system randomly assigns one of the three corresponding title templates. This random assignment introduces lexical and stylistic variation while preserving the underlying analytical intent of each sub-template.
\subsubsection{Table template design} 
\label{sec:Table_template_design}
Tables consist of a caption and a data body. In practical business analytics, the table caption typically specifies the analysis topic and data scope (e.g., location and time period in real-estate contexts), while the table body presents not raw data directly, but aggregated and statistically summarized results. For example, a real-estate company typically shows monthly or quarterly sales summaries across regions, rather than transaction records for each individual building.
Therefore, for each theme's particular statistical analysis requirements, we collaborate with real-estate domain experts to define the core statistical logic. For each theme $M_i$, we design 1–3 specific analytical dimensions and define 11 distinct statistical functions $F_k$. Each statistical function $F_k$ corresponds one-to-one to a specific table template. Intuitively, $SQL_k$ (defined below) specifies \emph{what subset of records} to retrieve, while $F_k$ encodes \emph{how to aggregate and summarize} the retrieved data.
During source slide generation, each sub-template $T_j$ within theme $M_i$ randomly samples a statistical function $F_k$ from the theme's function set, applies it to generate results, and renders the output using the corresponding table body template.
\paragraph{Caption generation.}
We construct ``caption + table body'' generation templates based on these statistical functions. The caption explicitly specifies the analytic scope and context of the table (e.g., temporal span or geographic region). We adopt variable-based templates and, for each statistical function, we design three caption templates $C_p$ (33 in total). As illustrated in Figure~\ref{fig:template}, each caption template combines fixed text (in black) with variable placeholders (in color).
To instantiate these templates, we construct a \textbf{Field Mapping Dictionary} (released with the dataset; see Table ~\ref{tab:field_mapping_dictionary}) that links template-level variables to concrete database fields. During the instantiation process, the system queries this dictionary to resolve all placeholders. By retrieving the corresponding concrete values (e.g., ``Beijing'') from the database, it transforms the abstract template into a final caption with explicit business semantics.
\paragraph{Table body generation.}
The table body is primarily responsible for presenting core data derived from aggregation and statistical analysis. Structurally, the table body template is composed of an SQL query template $SQL_k$ paired with a function template $F_k$, corresponding to the two primary generation stages: data retrieval and statistical computation.
First, to ensure the table content aligns strictly with the caption's scope, we leverage shared variables: we populate the SQL query template $SQL_k$ using the same variables resolved via the Field Mapping Dictionary during caption instantiation, thereby retrieving precise raw data. Subsequently, the system configures the computation logic by querying the \textbf{Parameter Candidate Dictionary} (released with the dataset; see Table ~\ref{tab:parameter_dictionary}) to randomly sample function-specific parameters (e.g., step sizes for binning) for $F_k$.
In parallel, we specifically decouple header generation to introduce linguistic diversity. The system accesses the \textbf{Header Alias Dictionary} (released with the dataset; see Table~\ref{tab:header_alias_dictionary})—which maps standard metrics to multiple synonymous variants—and randomly selects distinct aliases for both column and row headers. This mechanism ensures that tables presenting the same underlying data logic can exhibit varied terminological expressions.
Finally, the system passes these configuration inputs—structural parameters and selected headers—alongside the retrieved raw data into $F_k$ to generate the final table body.
\paragraph{Canonical table structures.}
For table presentation formats, we organize all computed results into three canonical table structures, each corresponding to distinct analytic perspectives:
\begin{itemize} \item \textbf{Field-constraint type:} rows correspond to statistical fields/metrics (e.g., average price, supply, sales), while columns correspond to a single constraint dimension (e.g., year or area bin), supporting multi-metric comparison under aligned conditions. \item \textbf{Constraint-field type:} rows correspond to constraint values, while columns correspond to statistical fields/metrics, suitable when the number of metrics is small and the focus is comparison across constraint values. \item \textbf{Cross-constraint type:} both rows and columns encode different constraints, with cell values reflecting their intersections; this structure most strongly tests system capability for understanding compound constraints. \end{itemize}
These three canonical structures(see Figure ~\ref{fig:table structure}) allow us to systematically probe models’ ability to interpret tables with simple constraints (Field-constraint and Constraint-field) and more complex, compositional constraints (Cross-constraint).

\begin{figure*}[t]
  \includegraphics[width=\linewidth]{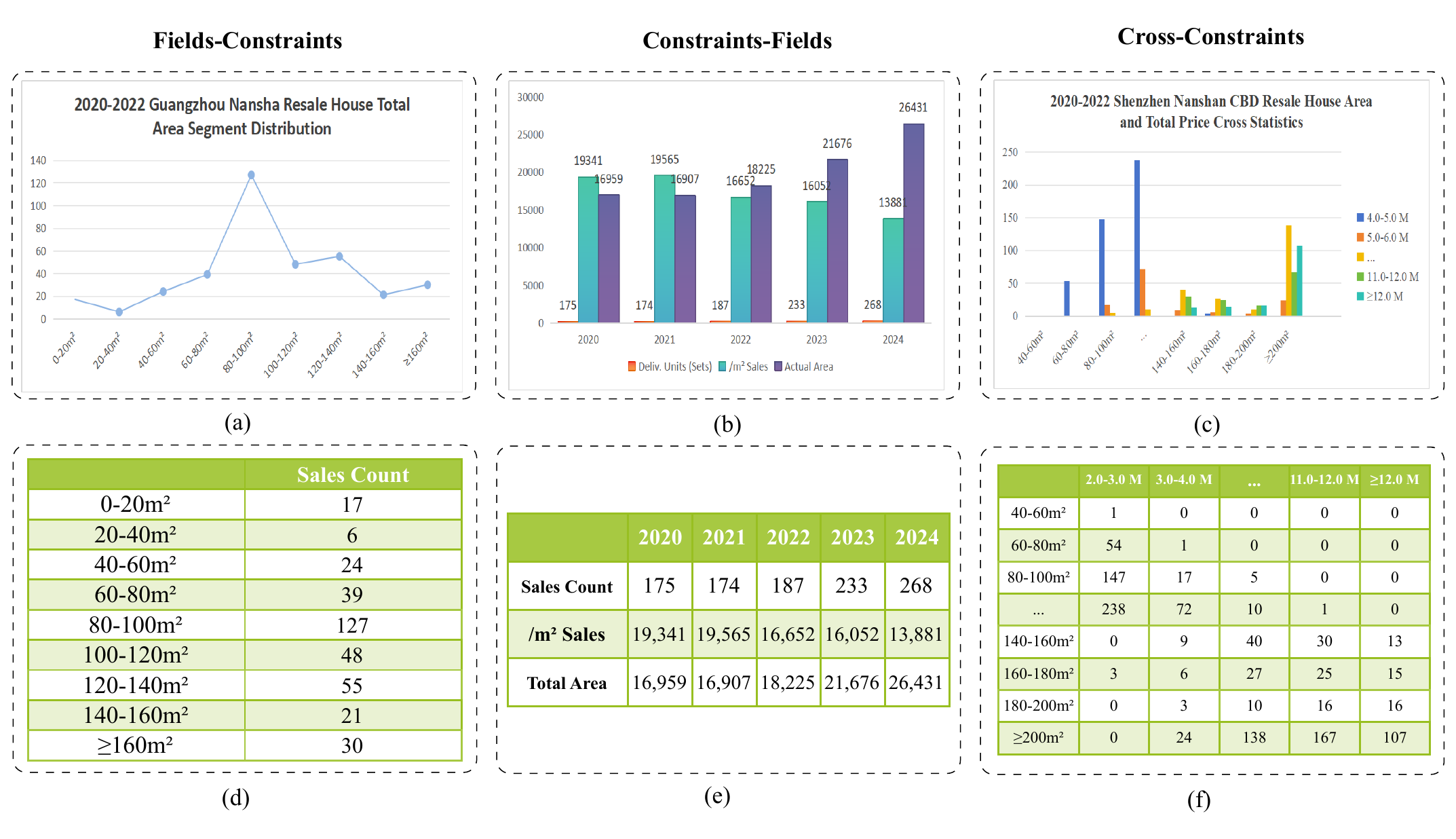}
  \caption{Illustration of three canonical table structures: Field-constraint, Constraint-field, and Cross-constraint.}
  \label{fig:table structure}
\end{figure*}

\subsubsection{Chart template design}
Charts are treated not as independent content units, but as visual projections of the same structured data used in tables. This approach allows for diverse presentation formats of the same underlying dataset. Consequently, the design process for chart templates mirrors that of table templates (Section~\ref{sec:Table template design}), strictly adhering to the principle of ``homologous data, heterogeneous presentation.''
Specifically, we reuse the statistical functions $F_k$ and SQL retrieval logic originally defined for tables to obtain the underlying data. We then leverage chart rendering tools to render this data according to the specified chart type. In the current release, we support bar and line charts, which cover the predominant chart types observed in our real-world reference reports.
\subsubsection{Summary template design}
The Summary synthesizes complex data into textual insights. Analyzing real-world reports reveals that summaries typically use stable phrasing structures with dynamically injected data conclusions. To automate this, we implemented dedicated insight-extraction logic for each of the 11 statistical functions. For each function, we automatically compute a set of key metrics (e.g., trends, growth rates, peak values)—typically 2–4 core metrics per function—from the computed table data.
We then designed 3 distinct summary templates $\textit{Summary}_q$ for each function (33 in total) to ensure expressive diversity. Similar to caption and table body generation, summary templates are instantiated by leveraging the Field Mapping Dictionary to resolve variable placeholders (e.g., resolving city to ``Beijing''), ensuring consistent data context across all three slide components. Across all business themes, these templates collectively cover approximately 30–40 distinct business indicator types (e.g., month-on-month growth in average price, distribution of unit prices across area bins; see Table ~\ref{tab:realestatevariables}).
Crucially, this unified approach guarantees semantic consistency across titles, tables, charts, and summaries, which is essential for the dynamic update task we target.
\subsection{Source slide generation}
\label{sec:Source slide generation}
To support systematic template instantiation and slide generation, we develop \textbf{EasySlide}, a toolkit built on \texttt{python-pptx} (see Table~\ref{tab:function_docs}). EasySlide provides simplified interfaces for common slide operations such as adding titles, inserting tables, and rendering charts, improving code maintainability and enabling efficient batch generation.
The source slide generation process proceeds through four key steps:
\begin{itemize} 
\item \emph{Template instantiation:} The system selects a sub-template based on the business theme and populates placeholders (e.g., {\text{city}}) with concrete parameters to generate specific titles and captions. \item \emph{Data generation:} Using these parameters, the system executes SQL queries to retrieve raw data, which is then processed by statistical functions to compute table values and extract key insights for summaries. \item \emph{Content rendering:} The EasySlide toolkit renders the generated titles, tables, charts, and textual summaries onto slides based on the computed data, ensuring consistency in layout and style. 
\item \emph{Metadata annotation:} For each slide, a YAML file is generated recording two essential components: \texttt{$slide_filters$} captures the data pipeline (database tables, SQL templates, statistical functions, filter parameters), while \texttt{$template_slide$} documents the slide layout (element types, positions, and content roles). Together, these structured annotations provide precise grounding for subsequent dynamic content updates. 
\end{itemize}
We release these YAML files alongside the slide decks, so that future work can directly leverage the structured annotations for training, evaluation, and tool-based execution. This systematic process yields 7{,}685 distinct source slides as the foundation for our dataset.

\subsection{User Instruction and Target Slide Generation}
\label{sec:User Instruction and Target Slide Generation}
Based on the source slides, we further generate user instructions and target slides to construct complete instruction–execution triples. Each triple consists of a \emph{source slide}, a \emph{user instruction}, and a \emph{target slide}. All triples share the same external PostgreSQL database described in Section~\ref{sec:Data Collection}, which serves as the underlying data source for both slide generation and instruction execution.
We design two typical update scenarios based on the metadata of source slides (such as data filtering conditions and statistical function configurations):
\begin{enumerate} \item \textbf{Basic Replacement Instructions:} Update only core variables (e.g., temporal spans, geographic regions) while maintaining the statistical functions and analytical dimensions. For example: Generate an analysis of \{city\} \{block\} from \{start year\} to \{end year\}.'' \item \textbf{Customized-Parameter Instructions:} Further modify constraint parameters (e.g., area segmentation, price granularity), triggering cascading updates across data queries and statistical computations. For example: Change the area segmentation to {area range} and price granularity to {price range}.'' \end{enumerate}
To support the diversity of these two scenarios, we construct approximately 24 instruction templates. To flexibly accommodate different parameter combinations, we introduce an instruction variable dictionary, stored as \texttt{query\_filters} in the YAML metadata, that records the variables to be modified in the instruction and their new values. After instantiation, we employ Qwen3-80B to paraphrase the generated instructions, ensuring linguistic diversity and naturalness. To ensure one-to-one correspondence among source slides, user instructions, target slides, and metadata, we design the following three-step process:
\paragraph{Instruction sampling and filling.}
For each source slide, the system randomly selects an instruction template. Based on the parameter dictionary, the system randomly samples new parameter values (such as new geographic regions or statistical configurations), fills them into placeholders to generate a natural language instruction, and records the same new values in \texttt{query\_filters} to keep the instruction text and executable parameters synchronized.
\paragraph{Data querying and recomputation.}
The system merges \texttt{slide\_filters} (the original data query configuration from the source slide's YAML file) with \texttt{query\_filters} (parameter changes from the instruction) to generate \texttt{update\_filters}. According to \texttt{update\_filters}, it then executes SQL queries to retrieve updated data and recomputes tables, charts, and textual insights in a single, consistent update pipeline.
\paragraph{Rendering and metadata recording.}
Using the EasySlide toolkit, the system renders the updated tables, charts, and textual summaries onto new slides. During this process, the system strictly preserves the layout, color scheme, and typography of the source slide to ensure continuity of user experience. Simultaneously, the system generates a corresponding YAML metadata file for the target slide, containing \texttt{query\_filters} (instruction\ parameters), \texttt{update\_filters} (merged filtering conditions), and \texttt{output\_slide} (target slide information), forming a complete, traceable record.
Together, the source slide, user instruction, target slide, shared PostgreSQL database, and associated YAML metadata form a fully specified instruction–execution instance.

\subsection{Dataset Statistics}
\label{sec:Dataset-Statistics}
DynaSlide contains 20{,}036 instruction–execution samples, each represented as a triple ((\emph{source slide}, \emph{user instruction}, \emph{target slide})). Crucially, all samples are grounded in the shared structured database of real-estate transaction records described in Section~\ref{sec:Data Collection}, which provides the external data context for both slide generation and instruction execution.
We split the dataset into train/validation/test sets with a 6:2:2 ratio (12{,}022 / 4{,}007 / 4{,}007) and enforce zero overlap of sub-templates (T\_j) across splits to evaluate generalization to unseen layouts. The dataset covers six analytical themes, 11 statistical functions, and two instruction scenarios (basic variable replacement vs.\ parameter customization). Each sample is accompanied by structured YAML metadata and executable data dependencies (e.g., SQL templates and function parameters), enabling both end-to-end and component-level evaluation. Table~\ref{tab:dataset_stats} provides detailed statistics.
We release (i) all source and target slide decks in PPTX format, (ii) a PostgreSQL database dump of the underlying transaction records, and (iii) YAML metadata files and template dictionaries (Field Mapping, Parameter Candidates, and Header Aliases), together with the EasySlide toolkit configuration. This complete release supports reproducible research on dynamic slide update and facilitates future model development, ablation studies, and fine-grained evaluation.

\section{Method Implementation Details}
\subsection{Prompts Used in SlideAgent}
\label{sec:appendix_prompts}
This section summarizes all prompts employed within the SlideAgent framework.

In the \textbf{Slide Parsing and Representation} stage, the Multimodal Layout Parsing module utilizes a structured prompt (Figure~\ref{fig:prompt_layout_parsing}) to guide the VLM in predicting semantic labels and bounding boxes. For \textbf{Logic Extraction}, we use distinct prompts based on the sub-task: the Data Source Extraction module employs a schema-aligned slot-filling prompt (Figure~\ref{fig:prompt_data_source}) to ground slide elements to database fields. The Function Logic Extraction module uses two separate prompts: a function-calling prompt for the closed-domain scenario (Figure~\ref{fig:prompt_function_logic_closed}) to identify predefined statistical functions, and a generic decomposition prompt for the open-domain scenario (Figure~\ref{fig:prompt_function_logic_open}) to reconstruct aggregation logic from atomic components.

In the \textbf{Instruction-Driven Update} stage, the User Instruction Parsing module uses a state-update prompt (Figure~\ref{fig:prompt_instruction_parsing}) to modify the parameter state based on user input while preserving the JSON schema. Subsequently, the SQL Generation module employs a structured prompt (Figure~\ref{fig:prompt_sql_generation}) to convert these updated parameters into executable database queries. Finally, the Summary Update module utilizes a fact-aware rewriting prompt (Figure~\ref{fig:prompt_summary_update}), which takes the original summary and pre/post-update data as input to generate factually consistent textual insights.

\subsection{Details of Structured Slide Representation}
\label{sec:Details of Structured Slide Representation}
We define the structured representation of a slide as a list of element objects, blending VLM semantic predictions with precise attributes from the PPTX file. Specifically, each element object contains the following fields: Each element is represented by five fields: \textbf{id}, a unique identifier within the slide; \textbf{type}, the underlying PPTX shape type (e.g., \texttt{textBox}); \textbf{role}, the semantic label predicted by the VLM and confirmed via IoU matching (e.g., \texttt{caption}, \texttt{title}); \textbf{text}, the textual content; and \textbf{layout}, the geometric attributes from PPTX metadata, including \texttt{x}, \texttt{y}, \texttt{width}, and \texttt{height}.



\subsection{Algorithm for Open-Domain Logic Reconstruction}
\label{sec:appendix_algorithm}
The process of the synthesize analytical table interface is outlined in Algorithm~\ref{alg:generic_logic}. Instead of hard-coding logic for diverse table types, we formulate table generation as a unified \textit{Split--Apply--Combine} workflow. The system first slices raw data based on constraints, groups records by row and column headers, applies the target aggregation function, and finally reshapes the output into the target layout.

\begin{table}[h]
\centering
\small
\renewcommand{\arraystretch}{1.2}
\begin{tabular}{l p{0.75\linewidth}}
\toprule
\textbf{Symbol} & \textbf{Definition} \\
\midrule
$D$ & Raw data set retrieved from the database, consisting of $N$ records $\{r_1, \dots, r_N\}$. \\
$\mathcal{L}$ & The logic specification tuple: $\mathcal{L} = (S, H, C, F, O)$. \\
\midrule
$S$ & \textbf{Structure Type}. Determines the matrix layout strategy. \\
& $\bullet$ \texttt{FC} (Field-Constraint): Metrics as rows, constraints (e.g., time) as columns. \\
& $\bullet$ \texttt{CF} (Constraint-Field): Constraints as rows, metrics as columns. \\
& $\bullet$ \texttt{XC} (Cross-Constraint): Intersection of two constraints (e.g., City $\times$ Year). \\
\midrule
$H$ & \textbf{Headers}. $H = (H_{row}, H_{col})$. Defines the grouping keys for rows and columns. \\
\midrule
$C$ & \textbf{Constraint Spec}. A 5-element list specifying slicing logic and formatting: \newline
$C = [k, T, v_{start}, v_{end}, v_{step}]$. \\
& $\bullet$ $k$: \textbf{Constraint Type} (e.g., \texttt{"year"} or \texttt{"price"}). \\
& $\bullet$ $T$: \textbf{Parameterized Template} for dynamic formatting (e.g., \texttt{"{}-{}M"} or \texttt{"-"}). \\
& $\bullet$ $v_{start}, v_{end}$: \textbf{Boundary Values} defining the closed interval (e.g., \texttt{"2020", "2023"}). \\
& $\bullet$ $v_{step}$: \textbf{Parameter Interval} or step size (e.g., \texttt{"1"} or \texttt{"price\_range\_size"}). \\
\midrule
$F$ & \textbf{Source Fields}. The database columns to be aggregated (e.g., \texttt{sales\_vol}). \\
$O$ & \textbf{Operations}. The aggregation functions applied to $F$ (e.g., $\textsc{Sum}, \textsc{Count}$). \\
\bottomrule
\end{tabular}
\caption{Unified Notation for Logic Reconstruction Components.}
\label{tab:notation}
\end{table}

\begin{algorithm}[h]
\small
\caption{Generic Logic Reconstruction Pipeline}
\label{alg:generic_logic}
\begin{algorithmic}[1]
\REQUIRE Raw Data $D$; Logic Tuple $\mathcal{L} = (S, H, C, F, O)$ defined in Table~\ref{tab:notation}.
\STATE \textbf{// Step 1: Constraint Application (Split)}

\IF{$T$ indicates Range Binning (e.g., \texttt{"-"} or \texttt{"{}-{}M"})} 
    \STATE \COMMENT{Apply binning with boundaries $[v_{start}, v_{end}]$ and step $v_{step}$}
    \STATE $D' \leftarrow \text{ApplyBinning}(D, \text{field}=k, \text{start}=v_{start}, \text{end}=v_{end}, \text{step}=v_{step}, \text{fmt}=T)$
\ELSE
    \STATE \COMMENT{Case: Standard Filtering (e.g., $k=\text{City}$, $v_{start}=\text{"Beijing"}$)}
    \STATE $D' \leftarrow \{ r \in D \mid \text{ApplyFilter}(r[k], \text{op}=T, \text{val}=v_{start}) \}$
\ENDIF

\STATE \textbf{// Step 2: Determine Grouping Keys}
\STATE Initialize grouping set $G \leftarrow \emptyset$
\IF{$S = \texttt{XC}$} \STATE $G \leftarrow H_{row} \cup H_{col}$ 
\ELSIF{$S = \texttt{CF}$} \STATE $G \leftarrow H_{row}$ 
\ELSIF{$S = \texttt{FC}$} \STATE $G \leftarrow H_{col}$ 
\ENDIF

\STATE \textbf{// Step 3: Aggregation (Apply)}
\STATE $A \leftarrow \text{GroupBy}(D', G).\text{Aggregate}(F, O)$
\STATE \textit{Note: $A$ contains grouping columns $G$ and computed metric columns.}

\STATE \textbf{// Step 4: Layout Reshaping (Combine)}
\IF{$S = \texttt{XC}$}
    \STATE \COMMENT{Pivot: Map $H_{row}$ to index, $H_{col}$ to columns, using the first metric}
    \STATE $T_{out} \leftarrow \text{PivotTable}(A, \text{index}=H_{row}, \text{columns}=H_{col}, \text{values}=\text{Col}(F, O))$
\ELSIF{$S = \texttt{FC}$}
    \STATE $T_{out} \leftarrow A^\top$ \COMMENT{Transpose: Flip metrics to rows to match template}
\ELSE
    \STATE $T_{out} \leftarrow A$ \COMMENT{Standard layout, no reshape needed}
\ENDIF

\RETURN $T_{out}$
\end{algorithmic}
\end{algorithm}

\section{Additional Experiments}
\label{sec:appendix_details}

\subsection{Experimental Setup Details}
\label{sec:implementation_details}
 We employ few-shot prompting with task-specific demonstrations (typically 6--8 examples) and deterministic decoding (temperature=0.0) to ensure reproducibility. 
 The system integrates \texttt{python-pptx} (v0.6.21) for slide manipulation and PostgreSQL (v14.5) for data management, orchestrated via LangGraph (v0.6.4). All experiments are conducted on NVIDIA H100 GPUs using the vLLM framework for high-throughput inference.

\subsection{Supplementary Generalization Evidence}
\label{sec:appendix_generalization}

We conduct two additional experiments to assess generalization. First, since evaluating closed-source models on the full benchmark is prohibitively expensive, we compare open-weight and closed-source models on 500 randomly sampled cases from the template-generated test set. Closed-source models significantly outperform open-weight ones, with GPT-5.4 achieving the highest task success rate at 78.01\% / 76.66\% under the closed/open settings, followed by Qwen-MAX at 75.43\% / 72.20\%. Among open-weight models, the ranking remains unchanged: GPT-OSS-120B remains the strongest at 67.50\% / 63.47\%, slightly outperforming Qwen3-80B at 66.18\% / 60.82\%. These results further support that end-to-end slide updating is highly capability-dependent, with stronger models consistently achieving better performance. Second, we compare model performance on human-authored slides (see Section~\ref{sec:Dataset-Statistics}) against the same 500 templated cases. Trends remain similar: GPT-OSS-120B achieves 67.71\% / 61.46\% and Qwen3-14B achieves 33.33\% / 25.00\% under the closed/open settings, preserving the same open-weight ranking as the benchmark test set. This consistency supports our assumption that report slide updates are primarily layout-preserving edits on element contents, suggesting our benchmark measures generalizable slide-updating capability rather than template-specific matching.

\begin{table}[t]
\centering
\small
\setlength{\tabcolsep}{5pt}
\renewcommand{\arraystretch}{1.15}
\resizebox{\columnwidth}{!}{%
\begin{tabular}{lcccc}
\toprule
& \multicolumn{2}{c}{\textbf{(A) 500 template-generated}} & \multicolumn{2}{c}{\textbf{(B) Human-authored}} \\
\cmidrule(lr){2-3}\cmidrule(lr){4-5}
\textbf{Model} & \textbf{Closed} & \textbf{Open} & \textbf{Closed} & \textbf{Open} \\
\midrule
GPT-OSS-120B & 67.50 & 63.47 & 67.71 & 61.46 \\
GPT-OSS-20B & 46.80 & 35.66 & 45.83 & 35.42 \\
Qwen3-80B & 66.18 & 60.82 & 64.58 & 59.38 \\
Qwen3-30B & 50.24 & 43.58 & 48.96 & 41.67 \\
Qwen3-14B & 35.71 & 27.34 & 33.33 & 25.00 \\
\midrule
Qwen-MAX & 75.43 & 72.20 & 73.96 & 69.79 \\
GPT-5.4 & 78.01 & 76.66 & 77.08 & 75.00 \\
\bottomrule
\end{tabular}
}
\caption{Supplementary generalization results (task success rate\%).}
\label{tab:supp-generalization}
\end{table}

\subsection{Case Study}
\label{sec:appendix_failure_cases}

We observe two representative failure modes in SlideAgent. The first is error accumulation: once an upstream module makes a mistake in slide parsing, instruction grounding, SQL generation, or tool execution, the error propagates through later stages and eventually leads to inconsistency among tables, charts, and textual conclusions. In this sense, a local mistake rarely remains local; it instead distorts the entire update chain. The second is the low precision of Summary Update: even when the preceding structured steps are correct, the final summary may still fail to express the updated results with sufficient numerical and semantic fidelity. This weakness reflects both the intrinsic difficulty of fact-aware text generation and the strictness of our evaluation criterion. We therefore use exact match for summary rewriting to enforce semantic and numerical correctness. For example, if second-hand housing sales in Liangxiang, Beijing increase from 836 to 1,355, the correct increase is about 62.1\%; a rewritten summary reporting 63.2\% is still counted as incorrect. These two failure modes show that the main challenge lies not only in executing each module correctly, but also in maintaining precise cross-module consistency through to the final narrative.

\begin{table*}[h]
  \centering
  \small
  \setlength{\tabcolsep}{6pt}
  \renewcommand{\arraystretch}{1.4} 
  \begin{tabular}{l l p{7.5cm}} 
    \toprule
    \textbf{Template Variable} & \textbf{Source Column} & \textbf{Description} \\ 
    \midrule
    \texttt{start\_year} & \texttt{date\_code} & Extracted year from the start date of the analysis period (e.g., 2023). \\ 
    \texttt{end\_year}   & \texttt{date\_code} & Extracted year from the end date of the analysis period (e.g., 2024). \\ 
    \texttt{month}       & \texttt{date\_code} & Truncated to month-level precision to represent the specific reference month (e.g., 2024-01). \\ 
    \cmidrule(l){1-3}
    \texttt{city}        & \texttt{city}       & Direct mapping to the target 
 city. \\ 
    \texttt{block}       & \texttt{block}      & Direct mapping to the administrative block. \\ 
    \texttt{project}     & \texttt{project}    & Direct mapping to the residential  project. \\ 
    \bottomrule
  \end{tabular}
\caption{\textbf{Field Mapping Dictionary}. This dictionary defines how template-level variables are derived from concrete database columns.}
\label{tab:field_mapping_dictionary}
\end{table*}

\begin{table*}[h]
  \centering
  \small
  \renewcommand{\arraystretch}{1.3}
  \setlength{\tabcolsep}{6pt}
  \begin{tabular}{l p{5.0cm} p{6.5cm}} 
    \toprule
    \textbf{Parameter Variable} & \textbf{Candidate Value Set} & \textbf{Description} \\ 
    \midrule
    \texttt{area\_bin\_step}    & \{10, 15, 20, 25, 30\}       & Sets the interval width ($m^2$) for grouping properties by floor area. \\ 
    \cmidrule(l){1-3} 
    \texttt{price\_bin\_step}   & \{0.75, 1.0, 1.5, 1.75, 2.0\} & Sets the interval width (Million CNY) for grouping properties by total price. \\ 
    \bottomrule
  \end{tabular}
\caption{\textbf{Parameter Candidate Dictionary}. This dictionary defines the valid search space for parameters variable used in statistical functions ($F_k$).}
  \label{tab:parameter_dictionary}
\end{table*}

\begin{table*}[h]
  \centering
  \small
  \renewcommand{\arraystretch}{1.4}
  \setlength{\tabcolsep}{5pt}
  
\begin{tabular}{l p{6.8cm} p{5.5cm}} 
    \toprule
    \textbf{Header} & \textbf{Alias Variants} (Professional Selection) & \textbf{Description} \\ 
    \midrule
    
    \texttt{supply volume} 
    & \{Supply Volume, New Listings, Listing Count, Market Supply, Total Listings, Supply (Units), New Supply\} 
    & Number of residential units listed for sale. \\ 
    \cmidrule(l){1-3}
    
    \texttt{trade volume} 
    & \{Trade Volume, Sales Volume, Transaction Count, Units Sold, Deal Count, Total Sales, Turnover Volume\} 
    & Number of residential units sold. \\ 
    \cmidrule(l){1-3}
    
    \texttt{area range counts} 
    & \{Volume by Area, Sales by Size, Count by Area, Area Segment Volume, Units by Size, Size Distribution\} 
    & Transaction count grouped by floor area segments. \\ 
    \cmidrule(l){1-3}
    
    \texttt{price range counts} 
    & \{Volume by Price, Sales by Price, Count by Price, Price Segment Volume, Units by Price, Price Distribution\} 
    & Transaction count grouped by total price intervals. \\ 
    \cmidrule(l){1-3}
    
    \texttt{total supply area} 
    & \{Total Supply Area, Supply Floor Area, Total Listing Area, Aggregate Supply ($m^2$), Supply Size, Listed Area\} 
    & Aggregate floor area of listed properties. \\ 
    \cmidrule(l){1-3}
    
    \texttt{total trade area} 
    & \{Total Trade Area, Total Sales Area, Transaction Area, Sold Floor Area, Aggregate Sales ($m^2$), Sold Area\} 
    & Aggregate floor area of sold properties. \\ 
    \cmidrule(l){1-3}
    
    \texttt{avg price} 
    & \{Avg. Total Price, Avg. House Price, Mean Transaction Price, Avg. Unit Cost, Per-Unit Price, Avg. Deal Price\} 
    & Average price per housing unit (not per sqm). \\ 
    \cmidrule(l){1-3}
    
    \texttt{unit price} 
    & \{Unit Price, Price per $m^2$, Avg. Price ($/m^2$), Sqm Price, Transaction Price ($/m^2$), Market Rate\} 
    & Average price per square meter. \\ 
    \cmidrule(l){1-3}
    
    \texttt{area\_range} 
    & \{Area Range, Floor Area ($m^2$), Size Band, Property Size, Area Segment, Unit Size\} 
    & Grouping dimension for floor area intervals. \\ 
    \cmidrule(l){1-3}
    
    \texttt{price\_range} 
    & \{Price Range, Total Price (Band), Price Segment, Value Tier, Budget Range, Price Bracket\} 
    & Grouping dimension for total price intervals. \\ 
    \bottomrule
  \end{tabular}
  \caption{\textbf{Header Alias Dictionary}. This dictionary defines synonymous aliases for column and row headers across different statistical analyses. Multiple variants of the same metric enable linguistic diversity in table presentations.}
    \label{tab:header_alias_dictionary}
\end{table*}

\begin{table*}[t]
\centering
\small
\caption{Definitions of real-estate market analysis variables. All variables are automatically calculated from the transaction database.}
\label{tab:realestatevariables}
{%
\renewcommand{\arraystretch}{1.3}
\setlength{\tabcolsep}{6pt}      

\begin{tabular}{p{0.32\textwidth} p{0.63\textwidth}}
\toprule 
\textbf{Variable Name} & \textbf{Description} \\
\midrule 
\texttt{City} & Name of the target city for analysis (e.g., Beijing). \\
\texttt{Block} & Name of the specific residential block or sub-district. \\
\texttt{Start\_Year} & The starting year of the analysis period (e.g., 2020). \\
\texttt{End\_Year} & The ending year of the analysis period (e.g., 2024). \\
\texttt{Start\_Month} & The starting month for monthly trend analysis. \\
\texttt{End\_Month} & The ending month for monthly trend analysis. \\

\texttt{Seg\_SupplyDemand\_Core\_Area} &
Core demand area segment: floor area range where primary housing demand is most concentrated, indicating the core supply--demand space. \\

\texttt{Seg\_SupplyDemand\_Upgrade\_Area} &
Upgrade demand area segment: floor area range for improved-living demand, indicating the upgraded housing segment. \\

\texttt{Total\_Transaction\_Units} &
Total transaction units during the observed period, reflecting overall market activity. \\

\texttt{Modal\_Price\_Segment} &
Dominant transaction price range with the highest volume, representing the prevailing market price level. \\

\texttt{Modal\_Area\_Segment} &
Dominant transaction area range with the highest volume, indicating mainstream housing size preferences. \\

\texttt{Peak\_Segment\_Volume} &
Peak cell transaction volume in the price--area matrix, highlighting the most popular specific product subtype. \\

\texttt{Dominant\_Area\_Segment} &
Label of the area segment with the highest transaction concentration (e.g., 80--100 m$^2$). \\

\texttt{Dominant\_Area\_Segment\_Volume} &
Total transaction volume within the dominant area segment. \\

\texttt{Dominant\_Price\_Segment} &
Label of the price segment with the highest transaction concentration. \\

\texttt{Dominant\_Price\_Segment\_Volume} &
Total transaction volume within the dominant price segment. \\

\texttt{Base\_Period\_Traded\_Area} &
Total annual traded area (m$^2$) in the base year. \\

\texttt{Terminal\_Period\_Traded\_Area} &
Total annual traded area (m$^2$) in the terminal year. \\

\texttt{Base\_Period\_Avg\_Price} &
Average transaction price (CNY/m$^2$) in the base year. \\

\texttt{Terminal\_Period\_Avg\_Price} &
Average transaction price (CNY/m$^2$) in the terminal year. \\

\texttt{Total\_Area\_Change\_Pct} &  
Percentage change of total traded area from base to terminal year (e.g., 0.25 = 25\%), indicating market expansion or contraction. \\

\texttt{Total\_Price\_Change\_Pct} & 
Percentage change in average transaction price from base to terminal year. \\

\texttt{Area\_Trend\_Direction} & 
Direction of traded area trend: ``Increasing'' or ``Decreasing''. \\

\texttt{Price\_Trend\_Direction} & 
Direction of price trend: ``Increasing'' or ``Decreasing''. \\

\texttt{Absolute\_Price\_Change} &
Absolute price change (CNY/m$^2$) between terminal and base periods. \\

\texttt{Supply\_Trend\_Direction} & 
Direction of supply trend: ``Increasing'' or ``Decreasing''. \\

\texttt{Transaction\_Trend\_Direction} & 
Direction of transaction volume trend: ``Increasing'' or ``Decreasing''. \\

\texttt{Base\_Period\_Supply\_Units} & 
Total number of supplied units in the base year. \\

\texttt{Terminal\_Period\_Supply\_Units} & 
Total number of supplied units in the terminal year. \\

\texttt{Base\_Period\_Transaction\_Units} & 
Total transaction units in the base year. \\

\texttt{Terminal\_Period\_Transaction\_Units} & 
Total transaction units in the terminal year. \\

\texttt{Total\_Supply\_Change\_Pct} & 
Percentage change in supply volume between base and terminal years. \\

\texttt{Total\_Transaction\_Change\_Pct} & 
Percentage change in total transactions between base and terminal years. \\

\texttt{Transaction\_Change\_Units} & 
Absolute change in transaction units (terminal minus base). \\

\texttt{Base\_Period\_Transaction\_Price} &
Transaction price level in the base year. \\

\texttt{Terminal\_Period\_Transaction\_Price} &
Transaction price level in the terminal year. \\

\texttt{Transaction\_Change\_Price} &
Absolute change in transaction price between base and terminal periods. \\

\bottomrule 
\end{tabular}%
}
\end{table*}

\begin{table*}[t]
  \centering
  \small
  \setlength{\tabcolsep}{8pt}
  \renewcommand{\arraystretch}{1.5}
  \begin{tabular}{p{3.5cm} p{11.5cm}}
    \toprule
    \textbf{Function} & \textbf{Description} \\
    \midrule
    \texttt{create\_slide} & Initializes a new slide page within the presentation deck. This function acts as the container for subsequent elements, allowing users to specify an optional slide layout template (e.g., title-only, blank). \\
    \midrule
    \texttt{add\_title} & Adds a prominent title element to the current slide context. This is typically used to define the primary topic or section header for the slide. \\
    \midrule
    \texttt{add\_text} & Inserts a text box containing specified content at given coordinates. This function supports multi-line text, bullet points, and basic rich text formatting for detailed descriptions. \\
    \midrule
    \texttt{add\_table} & Renders a structured data table on the slide. The function automatically calculates and adjusts cell dimensions (row heights and column widths) to fit the provided dataset visually. \\
    \midrule
    \texttt{add\_line\_chart} & Generates a line chart visualization to display trends over time or continuous variables. It supports plotting multiple data series for comparative analysis. \\
    \midrule
    \texttt{add\_bar\_chart} & Creates a bar chart to compare quantitative values across distinct categories. Users can specify orientation (vertical or horizontal) to best fit the data distribution. \\
    \bottomrule
  \end{tabular}
  \caption{Description of core functions for slide generation.}
    \label{tab:function_docs}
\end{table*}

\begin{table*}[t]
  \centering
  \small
  \renewcommand{\arraystretch}{1.0}
  \setlength{\tabcolsep}{5pt}
  \begin{tabular}{p{3.8cm} p{4.2cm} r}
    \toprule
    \textbf{Category} & \textbf{Dimension} & \textbf{Quantity} \\
    \midrule
    \multirow{4}{=}{\textbf{Dataset Splits}} & Total Samples & 20,036 \\
    & Train Set & 12,022 \\
    & Validation Set & 4,007 \\
    & Test Set & 4,007 \\
    \midrule
    \multirow{5}{=}{\textbf{Template\\Statistics}} & Analysis Themes & 6 \\
    & Statistical Operations & 11 \\
    & Template Types & 34 \\
    & Avg. Elements per Slide & 4.2 \\
    & Source Slides (Seeds) & 7,685 \\
    \midrule
    \multirow{4}{=}{\textbf{Instruction\\Statistics}} & Basic Instructions & 13,016 \\
    & Customized Instructions & 7,020 \\
    & Avg. Instruction Length & 20.11 \\
    & Avg. Parameters per Instruction & 3.8 \\
    \midrule
    \multirow{8}{=}{\textbf{Annotation\\Components}} & Natural Language Instructions & 20,036 \\
    & Slide Titles and Captions & 45,854 \\
    & SQL Queries & 20,036 \\
    & Summaries & 20,036 \\
    & Statistical Tables (Total) & 25,041 \\
    & \quad Base Table & 4,919 \\
    & \quad Line Chart Table & 9,746 \\
    & \quad Bar Chart Table & 10,376 \\
    \bottomrule
  \end{tabular}
\caption{Detailed statistics of the DynaSlide dataset. The dataset, derived from \textbf{7,685 source slides}, covers four core elements (titles, tables, charts, summaries) and includes full annotations for SQL, logic, and visual layouts.}
  \label{tab:dataset_stats}
\end{table*}

\onecolumn
\setlength{\tabcolsep}{4pt}      
\renewcommand{\arraystretch}{1.2} 

\begin{longtable}{L{2.4cm} C{2.2cm} L{4.8cm} L{6.0cm}}
\caption{Full Template Specification for Real-Estate Market Text Generation. Variables are denoted in \texttt{monospace} and are automatically populated during generation.} 
\label{tab:full_templates} \\
\toprule
\textbf{Title Template} & \textbf{Function} & \textbf{Caption Template} & \textbf{Summary Template} \\
\midrule
\endfirsthead

\multicolumn{4}{c}{{\bfseries \tablename\ \thetable{} -- continued from previous page}} \\
\toprule
\textbf{Title Template} & \textbf{Function} & \textbf{Caption Template} & \textbf{Summary Template} \\
\midrule
\endhead

\midrule
\multicolumn{4}{r}{\textit{Continued on next page...}} \\
\bottomrule
\endfoot

\bottomrule
\endlastfoot


1. Block Area Segment Distribution \par\smallskip
2. Analysis of Block Area Division \par\smallskip
3. Distribution of Block Area Segments & 
Supply-Transaction Unit Statistics & 
1. \VAR{{Start\_Year}}-\VAR{{End\_Year}} Supply and Transaction Unit Statistics in \VAR{{City}}'s \VAR{{Block}} \par\smallskip
2. \VAR{City} \VAR{Block}: Supply \& Sales Volume Analysis, \VAR{Start\_Year}-\VAR{End\_Year} \par\smallskip
3. Analysis of Property Units Supplied vs. Sold in \VAR{City}'s \VAR{Block} (\VAR{Start\_Year}-\VAR{End\_Year})& 
1. From \VAR{{Start\_Year}} to \VAR{{End\_Year}}, \VAR{{Block}}'s core supply-demand area was \VAR{{Seg_SupplyDemand_Core_Area}} m\textsuperscript{2}, with the upgrade-oriented segment centered on \VAR{{Seg_SupplyDemand_Upgrade_Area}} m\textsuperscript{2}. \par\smallskip
2. Between \VAR{{Start\_Year}} and \VAR{{End\_Year}}, the market structure in \VAR{{Block}} was defined by a core demand range of \VAR{{Seg_SupplyDemand_Core_Area}} m\textsuperscript{2} and an upgrade tier of \VAR{{Seg_SupplyDemand_Upgrade_Area}} m\textsuperscript{2}. \par\smallskip
3. The \VAR{{Block}} sector exhibited a dual-tier segmentation from \VAR{{Start\_Year}}-\VAR{{End\_Year}}: a primary volume cluster at \VAR{{Seg_SupplyDemand_Core_Area}} m\textsuperscript{2} and a secondary upgrade cluster at \VAR{{Seg_SupplyDemand_Upgrade_Area}} m\textsuperscript{2}. \\
\midrule

1. New-House Cross-Structure Analysis \par\smallskip
2. New Residential Portfolio Composition  \par\smallskip
3. New Construction Inventory Structure Analysis  &  
Area $\times$ Price Cross Pivot & 
1. \VAR{{Start\_Year}}-\VAR{{End\_Year}} \VAR{{City}} \VAR{{Block}} Area and Total Price Cross Statistics \par\smallskip
2. Cross-Analysis of Property Size and Price Points in \VAR{City}'s \VAR{Block} (\VAR{Start\_Year}-\VAR{End\_Year}) \par\smallskip
3. \VAR{City} \VAR{Block}: Correlation between Unit Area and Total Price (\VAR{Start\_Year}-\VAR{End\_Year})
& 
1. From \VAR{{Start\_Year}} to \VAR{{End\_Year}}, a total of \VAR{{Total_Transaction_Units}} units were transacted, with the \VAR{{Modal_Price_Segment}} price segment and \VAR{{Modal_Area_Segment}} area segment having the highest transactions at \VAR{{Peak_Segment_Volume}} units. \par\smallskip
2. Out of \VAR{{Total_Transaction_Units}} total transactions during \VAR{{Start\_Year}}-\VAR{{End\_Year}}, the peak velocity of \VAR{{Peak_Segment_Volume}} units occurred at the intersection of the \VAR{{Modal_Price_Segment}} price band and \VAR{{Modal_Area_Segment}} area band. \par\smallskip
3. The period \VAR{{Start\_Year}}-\VAR{{End\_Year}} saw \VAR{{Total_Transaction_Units}} total sales; the most active cross-segment was \VAR{{Modal_Price_Segment}} combined with \VAR{{Modal_Area_Segment}}, contributing \VAR{{Peak_Segment_Volume}} units. \\
\midrule

1. New-House Cross-Structure Analysis \par\smallskip
2. New Residential Portfolio Composition  \par\smallskip
3. New Construction Inventory Structure Analysis  & 
Area Segment Distribution & 
1.\VAR{{Start\_Year}}-\VAR{{End\_Year}} \VAR{{City}} \VAR{{Block}} Total Area Segment Distribution Statistics \par\smallskip
2. Distribution of Transactions by Property Size Segment in \VAR{City}'s \VAR{Block} (\VAR{Start\_Year}-\VAR{End\_Year}) \par\smallskip
3. \VAR{City} \VAR{Block}: Analysis of Market Share by Unit Area Brackets (\VAR{Start\_Year}-\VAR{End\_Year})
& 
1. Mainstream types concentrate in \VAR{{Dominant_Area_Segment}} segments, totaling \VAR{{Dominant_Area_Segment_Volume}} units. \par\smallskip
2. A volume of \VAR{{Dominant_Area_Segment_Volume}} units indicates that the \VAR{{Dominant_Area_Segment}} range represents the dominant area concentration. \par\smallskip
3. The \VAR{{Dominant_Area_Segment}} typology emerged as the mainstream segment, amassing a total of \VAR{{Dominant_Area_Segment_Volume}} units. \\
\midrule

1. New-House Cross-Structure Analysis \par\smallskip
2. New Residential Portfolio Composition  \par\smallskip
3. New Construction Inventory Structure Analysis  & 
Price Segment Distribution & 
1. \VAR{{Start\_Year}}-\VAR{{End\_Year}} \VAR{{City}} \VAR{{Block}} Total Price Segment Distribution Statistics \par\smallskip
2. Distribution of Transactions by Price Point Segment in \VAR{City}'s \VAR{Block} (\VAR{Start\_Year}-\VAR{End\_Year}) \par\smallskip
3. Sales Breakdown by Price Range Categories for \VAR{City}'s \VAR{Block}, \VAR{Start\_Year}-\VAR{End\_Year}
& 
1. Mainstream types concentrate in \VAR{{Dominant_Price_Segment}} segments, totaling \VAR{{Dominant_Price_Segment_Volume}} units. \par\smallskip
2. The \VAR{{Dominant_Price_Segment}} price bracket captured the majority of interest, accumulating \VAR{{Dominant_Price_Segment_Volume}} units. \par\smallskip
3. With \VAR{{Dominant_Price_Segment_Volume}} units, the \VAR{{Dominant_Price_Segment}} segment constitutes the primary price concentration for the sector. \\
\midrule

1. Resale-House Cross-Structure Analysis \par\smallskip
2. Resale Residential Portfolio Assessment  \par\smallskip
3. Secondary Market Inventory Structure Study  &
Area $\times$ Price Cross Pivot & 
1. \VAR{{Start\_Year}}-\VAR{{End\_Year}} \VAR{{City}} \VAR{{Block}} Resale House Area and Total Price Cross Statistics  \par\smallskip
2. Resale Market: Cross-Analysis of Property Size and Price in \VAR{City}'s \VAR{Block} (\VAR{Start\_Year}-\VAR{End\_Year}) \par\smallskip
3. Statistical Profile of Resale Homes by Area vs. Total Price in \VAR{City}'s \VAR{Block} (\VAR{Start\_Year}-\VAR{End\_Year})
& 
1. From \VAR{{Start\_Year}} to \VAR{{End\_Year}}, a total of \VAR{{Total_Transaction_Units}} units were transacted, with the \VAR{{Modal_Price_Segment}} price segment and \VAR{{Modal_Area_Segment}} area segment having the highest transactions at \VAR{{Peak_Segment_Volume}} units. \par\smallskip
2. Resale activity for \VAR{{Start\_Year}}-\VAR{{End\_Year}} totaled \VAR{{Total_Transaction_Units}} units, peaked by \VAR{{Peak_Segment_Volume}} sales in the \VAR{{Modal_Price_Segment}} / \VAR{{Modal_Area_Segment}} cross-segment. \par\smallskip
3. The \VAR{{Modal_Price_Segment}} and \VAR{{Modal_Area_Segment}} cohorts led the resale market with \VAR{{Peak_Segment_Volume}} units, driving a cumulative volume of \VAR{{Total_Transaction_Units}}. \\
\midrule

1. Resale-House Cross-Structure Analysis \par\smallskip
2. Resale Residential Portfolio Assessment  \par\smallskip
3. Secondary Market Inventory Structure Study  &
Area Segment Distribution & 
1. \VAR{{Start\_Year}}-\VAR{{End\_Year}} \VAR{{City}} \VAR{{Block}} Resale House Total Area Segment Distribution Statistics \par\smallskip
2. Resale Market Transaction Distribution by Property Size in \VAR{City}'s \VAR{Block} (\VAR{Start\_Year}-\VAR{End\_Year}) \par\smallskip
3. Breakdown of Existing Home Sales by Size Category in \VAR{City}'s \VAR{Block}, \VAR{Start\_Year}-\VAR{End\_Year}
& 
1. Mainstream types concentrate in the \VAR{{Dominant_Area_Segment}} segments, totaling \VAR{{Dominant_Area_Segment_Volume}} units. \par\smallskip
2. The resale inventory is heavily weighted in the \VAR{{Dominant_Area_Segment}} range, which accounts for \VAR{{Dominant_Area_Segment_Volume}} units. \par\smallskip
3. Accounting for \VAR{{Dominant_Area_Segment_Volume}} units, the \VAR{{Dominant_Area_Segment}} category stands out as the primary resale typology. \\
\midrule

1. Resale-House Cross-Structure Analysis \par\smallskip
2. Resale Residential Portfolio Assessment  \par\smallskip
3. Secondary Market Inventory Structure Study  &
Price Segment Distribution & 
1. \VAR{{Start\_Year}}-\VAR{{End\_Year}} \VAR{{City}} \VAR{{Block}} Resale House Total Price Segment Distribution Statistics \par\smallskip
2. Resale Market Transaction Distribution by Price Point in \VAR{City}'s \VAR{Block} (\VAR{Start\_Year}-\VAR{End\_Year}) \par\smallskip
3. Breakdown of Existing Home Sales by Price Range in \VAR{City}'s \VAR{Block}, \VAR{Start\_Year}-\VAR{End\_Year}
& 
1. Mainstream types concentrate in the \VAR{{Dominant_Price_Segment}} segments, totaling \VAR{{Dominant_Price_Segment_Volume}} units. \par\smallskip
2. The \VAR{{Dominant_Price_Segment}} price tier represents the core resale market, comprising \VAR{{Dominant_Price_Segment_Volume}} units. \par\smallskip
3. A total of \VAR{{Dominant_Price_Segment_Volume}} resale units clustered within the \VAR{{Dominant_Price_Segment_Volume}} price band. \\
\midrule

1. New-House Market Capacity Analysis \par\smallskip
2. New Construction Volume \& Supply Capacity   \par\smallskip
3. Emerging Residential Market Scale Evaluation  & 
Historical Capacity Summary &
1. \VAR{{City}} \VAR{{Block}} Historical Capacity Summary Statistics (\VAR{{Start\_Year}}-\VAR{{End\_Year}}) \par\smallskip
2. Historical Market Volume Summary for \VAR{City}'s \VAR{Block} (\VAR{Start\_Year}-\VAR{End\_Year}) \par\smallskip
3. Summary of Past Market Scale Statistics for \VAR{City}'s \VAR{Block}, \VAR{Start\_Year}-\VAR{End\_Year}
& 
1. From \VAR{{Start\_Year}} to \VAR{{End\_Year}}, \VAR{{Block}}'s traded area \VAR{{Area_Trend_Direction}} \VAR{{Total_Area_Change_Pct}}\% from \VAR{{Base_Period_Traded_Area}} m\textsuperscript{2} to \VAR{{Terminal_Period_Traded_Area}} m\textsuperscript{2}, and the average valuation also \VAR{{Price_Trend_Direction}} \VAR{{Total_Price_Change_Pct}}\% from \VAR{{Base_Period_Avg_Price}} to \VAR{{Terminal_Period_Avg_Price}} yuan/m\textsuperscript{2}. \par\smallskip

2. Over the \VAR{{Start\_Year}}-\VAR{{End\_Year}} period, the sector saw traded area \VAR{{Area_Trend_Direction}} by \VAR{{Total_Area_Change_Pct}}\% (moving from \VAR{{Base_Period_Traded_Area}} to \VAR{{Terminal_Period_Traded_Area}} m\textsuperscript{2}), while valuations \VAR{{Price_Trend_Direction}} by \VAR{{Total_Price_Change_Pct}} yuan/m\textsuperscript{2}. \par\smallskip

3. Starting at \VAR{Base_Period_Traded_Area} m\textsuperscript{2} 
and \VAR{Base_Period_Avg_Price} yuan/m\textsuperscript{2}, 
the market \VAR{Area_Trend_Direction} to \VAR{Terminal_Period_Traded_Area} m\textsuperscript{2} 
and \VAR{Terminal_Period_Avg_Price} yuan/m\textsuperscript{2} respectively, 
marking a volume shift of \VAR{Total_Area_Change_Pct}\% 
and a price shift of \VAR{Absolute_Price_Change} yuan/m\textsuperscript{2}.\\
\midrule

1. New-House Market Capacity Analysis \par\smallskip
2. New Construction Volume \& Supply Capacity   \par\smallskip
3. Emerging Residential Market Scale Evaluation  &
Annual Supply-Demand Comparison &

1. \VAR{{City}} \VAR{{Block}} Annual Supply-Demand Comparison Analysis (\VAR{{Start\_Year}}-\VAR{{End\_Year}})  \par\smallskip
2. \VAR{City} \VAR{Block}: Annual Comparison of Market Supply and Transaction Volume (\VAR{Start\_Year}-\VAR{End\_Year}) \par\smallskip
3. Analysis of Annual Supply-Demand Balance in \VAR{City}'s \VAR{Block} (\VAR{Start\_Year}-\VAR{End\_Year})
& 
1. From \VAR{{Start\_Year}} to \VAR{{End\_Year}}, new listings in this sector \VAR{{Supply_Trend_Direction}} from \VAR{{Base_Period_Supply_Units}} units to \VAR{{Terminal_Period_Supply_Units}} units (a \VAR{{Supply_Trend_Direction}} of \VAR{{Total_Supply_Change_Pct}}\%), and transaction volume \VAR{{Transaction_Trend_Direction}} from \VAR{{Base_Period_Transaction_Units}} units to \VAR{{Terminal_Period_Transaction_Units}} units (a \VAR{{Transaction_Trend_Direction}} of \VAR{{Total_Transaction_Change_Pct}}\%). \par\smallskip

2. While listings \VAR{{Supply_Trend_Direction}} by \VAR{{Total_Supply_Change_Pct}}\% (reaching \VAR{{Terminal_Period_Supply_Units}} units), transactions simultaneously \VAR{{Transaction_Trend_Direction}} by \VAR{{Total_Transaction_Change_Pct}}\% (ending at \VAR{{Terminal_Period_Transaction_Units}} units) between \VAR{{Start\_Year}} and \VAR{{End\_Year}}. \par\smallskip

3. Comparing \VAR{{Start\_Year}} to \VAR{{End\_Year}}, supply \VAR{{Supply_Trend_Direction}} to \VAR{{Terminal_Period_Supply_Units}} (\VAR{{Total_Supply_Change_Pct}}\% \VAR{{Supply_Trend_Direction}}), and demand \VAR{{Transaction_Trend_Direction}} to \VAR{{Terminal_Period_Transaction_Units}} (\VAR{{Total_Transaction_Change_Pct}}\% \VAR{{Transaction_Trend_Direction}}). \\
\midrule

1. New-House Market Capacity Analysis \par\smallskip
2. New Construction Volume \& Supply Capacity   \par\smallskip
3. Emerging Residential Market Scale Evaluation  &
Supply-Transaction Area & 

1. \VAR{{City}} \VAR{{Block}} : Historical Supply and Transaction Area Statistics (\VAR{{Start\_Year}}-\VAR{{End\_Year}})\par\smallskip
2. Statistical Review of Historical Supply and Transaction Area for \VAR{City}'s \VAR{Block} (\VAR{Start\_Year}-\VAR{End\_Year})\par\smallskip
3. Historical Data: Supplied vs. Sold Area in \VAR{City}'s \VAR{Block} (\VAR{Start\_Year}-\VAR{End\_Year})

& 
1. From \VAR{{Start\_Year}} to \VAR{{End\_Year}}, inventory in this region \VAR{{Supply_Trend_Direction}} by \VAR{{Total_Supply_Change_Pct}}\%, while the transaction area \VAR{{Transaction_Trend_Direction}} by \VAR{{Total_Transaction_Change_Pct}}\%. \par\smallskip
2. Area-wise inventory \VAR{{Supply_Trend_Direction}} at a \VAR{{Total_Supply_Change_Pct}}\% rate, contrasting with the transaction area which \VAR{{Transaction_Trend_Direction}} by \VAR{{Total_Transaction_Change_Pct}}\% through \VAR{{End\_Year}}. \par\smallskip
3. The region experienced a \VAR{{Total_Supply_Change_Pct}}\% \VAR{{Supply_Trend_Direction}} in supply area and a \VAR{{Total_Transaction_Change_Pct}}\% \VAR{{Transaction_Trend_Direction}} in sold area between \VAR{{Start\_Year}} and \VAR{{End\_Year}}. \\
\midrule

1. Resale-House Capacity \& Structure \par\smallskip
2. Resale Market Scale \& Product Breakdown   \par\smallskip
3. Secondary Market Stock \& Unit Composition  &
Historical Delivery Metrics & 
1. \VAR{{City}} \VAR{{Block}} Resale House: Overview of Historical Delivery Metrics (\VAR{{Start\_Year}}-\VAR{{End\_Year}}) \par\smallskip
2. Resale Market in \VAR{City}'s \VAR{Block}: A Look at Historical Transaction Metrics (\VAR{Start\_Year}-\VAR{End\_Year}) \par\smallskip
3. Historical Performance Overview for the Resale Market in \VAR{City}'s \VAR{Block} (\VAR{Start\_Year}-\VAR{End\_Year})

& 
1. From \VAR{{Start\_Year}} to \VAR{{End\_Year}}, the number of resale transactions in this area \VAR{{Transaction_Trend_Direction}} from \VAR{{Base_Period_Transaction_Units}} units to \VAR{{Terminal_Period_Transaction_Units}} units, with a cumulative \VAR{{Transaction_Trend_Direction}} of \VAR{{Transaction_Change_Units}} units, representing a \VAR{{Transaction_Trend_Direction}} of \VAR{{Total_Transaction_Change_Pct}}\%. \par\smallskip
2. Resale volumes moved from \VAR{{Base_Period_Transaction_Units}} to \VAR{{Terminal_Period_Transaction_Units}} units (\VAR{{Start\_Year}}-\VAR{{End\_Year}}), marking a \VAR{{Transaction_Change_Units}}-unit \VAR{{Transaction_Trend_Direction}} or a \VAR{{Total_Transaction_Change_Pct}}\% change. \par\smallskip
3. The market observed a \VAR{{Transaction_Trend_Direction}} in resales to \VAR{{Transaction_Change_Units}} units, a \VAR{{Total_Transaction_Change_Pct}}\% \VAR{{Transaction_Trend_Direction}} relative to the \VAR{{Base_Period_Transaction_Units}}-unit baseline in \VAR{{Start\_Year}}. \\
\midrule

1. Resale-House Capacity \& Structure \par\smallskip
2. Resale Market Scale \& Product Breakdown   \par\smallskip
3. Secondary Market Stock \& Unit Composition  &
Annual Delivery Unit Count & 

1. Historical Delivery Unit Count Statistics for \VAR{{City}} \VAR{{Block}} Resale House (\VAR{{Start\_Year}}-\VAR{{End\_Year}}) \par\smallskip
2. Historical Transaction Volume (Units) for Resale Homes in \VAR{City}'s \VAR{Block} (\VAR{Start\_Year}-\VAR{End\_Year}) \par\smallskip
3. Unit Sales Statistics for the Existing Home Market in \VAR{City}'s \VAR{Block} (\VAR{Start\_Year}-\VAR{End\_Year})
& 
1. From \VAR{{Start\_Year}} to \VAR{{End\_Year}}, the number of resale house transactions in this area \VAR{{Transaction_Trend_Direction}} from \VAR{{Base_Period_Transaction_Units}} units to \VAR{{Terminal_Period_Transaction_Units}} units, with a cumulative \VAR{{Transaction_Trend_Direction}} of \VAR{{Transaction_Change_Units}} units. \par\smallskip

2. A net \VAR{{Transaction_Trend_Direction}} of \VAR{{Transaction_Change_Units}} units was recorded as resale volume \VAR{{Transaction_Trend_Direction}} from \VAR{{Base_Period_Transaction_Units}} to \VAR{{Terminal_Period_Transaction_Units}} between \VAR{{Start\_Year}} and \VAR{{End\_Year}}. \par\smallskip

3. The trend \VAR{{Transaction_Trend_Direction}} brought resale figures to \VAR{{Terminal_Period_Transaction_Units}} by \VAR{{End\_Year}}, a \VAR{{Transaction_Trend_Direction}} of \VAR{{Transaction_Change_Units}} from the start of the period. \\
\midrule

1. Resale-House Capacity \& Structure \par\smallskip
2. Resale Market Scale \& Product Breakdown   \par\smallskip
3. Secondary Market Stock \& Unit Composition  &
Annual Average Price Trend & 

1.\VAR{{City}} \VAR{{Block}} Resale House: \VAR{{Start\_Year}}-\VAR{{End\_Year}} Annual Average Price Trend Analysis \par\smallskip
2. Annual Average Price Evolution for Resale Homes in \VAR{City}'s \VAR{Block} (\VAR{Start\_Year}-\VAR{End\_Year}) \par\smallskip
3.Analysis of Annual Mean Price Fluctuation for Resale Properties in \VAR{City}'s \VAR{Block}, \VAR{Start\_Year}-\VAR{End\_Year}
& 
1. Resale average prices trended \VAR{{Price_Trend_Direction}} from \VAR{{Base_Period_Transaction_Price}} yuan/m\textsuperscript{2} in \VAR{{Start\_Year}} to \VAR{{Terminal_Period_Transaction_Price}} yuan/m\textsuperscript{2} in \VAR{{End\_Year}} (\VAR{{Transaction_Change_Price}}). \par\smallskip

2. The market valuation followed a \VAR{{Price_Trend_Direction}} path, adjusting from \VAR{{Base_Period_Transaction_Price}} to \VAR{{Terminal_Period_Transaction_Price}} yuan/m\textsuperscript{2} amid \VAR{{Transaction_Change_Price}} conditions. \par\smallskip

3. Between \VAR{{Start\_Year}} and \VAR{{End\_Year}}, prices \VAR{{Price_Trend_Direction}} to \VAR{{Terminal_Period_Transaction_Price}} yuan/m\textsuperscript{2} from a base of \VAR{{Base_Period_Transaction_Price}}, characterizing a \VAR{{Transaction_Change_Price}} market. \\
\midrule

1. Neighborhood Price Trend \par\smallskip
2. Regional Price Movement \& Trajectory  \par\smallskip
3. Community Value Trend Analysis  &
Monthly Average Price Trend & 
1. \VAR{{City}} \VAR{{Block}} Monthly Average Price Trend Analysis, \VAR{Start\_Month}-\VAR{End\_Month}  \par\smallskip
2. Analysis of Monthly Average Price Movements in \VAR{City}'s \VAR{Block} (\VAR{Start\_Month}-\VAR{End\_Month}) \par\smallskip
3. Month-over-Month Average Price Fluctuation for \VAR{City}'s \VAR{Block} (\VAR{Start\_Month}-\VAR{End\_Month})
& 
1. Over \VAR{{Start_Month}}-\VAR{{End_Month}}, the average transaction price of apartments in this community went from \VAR{{Base_Period_Transaction_Price}} yuan/m\textsuperscript{2} \VAR{{Transaction_Trend_Direction}} to \VAR{{Terminal_Period_Transaction_Price}} yuan/m\textsuperscript{2}, with a cumulative \VAR{{Price_Trend_Direction}} of \VAR{{Transaction_Change_Price}} yuan/m\textsuperscript{2}. \par\smallskip
2. Between \VAR{{Start_Month}} and \VAR{{End_Month}}, prices \VAR{{Transaction_Trend_Direction}} by \VAR{{Transaction_Change_Price}} yuan/m\textsuperscript{2}, shifting the average from \VAR{{Base_Period_Transaction_Price}} to \VAR{{Terminal_Period_Transaction_Price}} yuan/m\textsuperscript{2}. \par\smallskip
3. A net \VAR{{Transaction_Trend_Direction}} of \VAR{{Transaction_Change_Price }} yuan/m\textsuperscript{2} was observed Over \VAR{{Start_Month}}-\VAR{{End_Month}}, as prices moved to \VAR{{Terminal_Period_Transaction_Price}} yuan/m\textsuperscript{2} from \VAR{{Base_Period_Transaction_Price}}. \\

\end{longtable}

\begin{figure}[t]
  \includegraphics[width=\columnwidth]{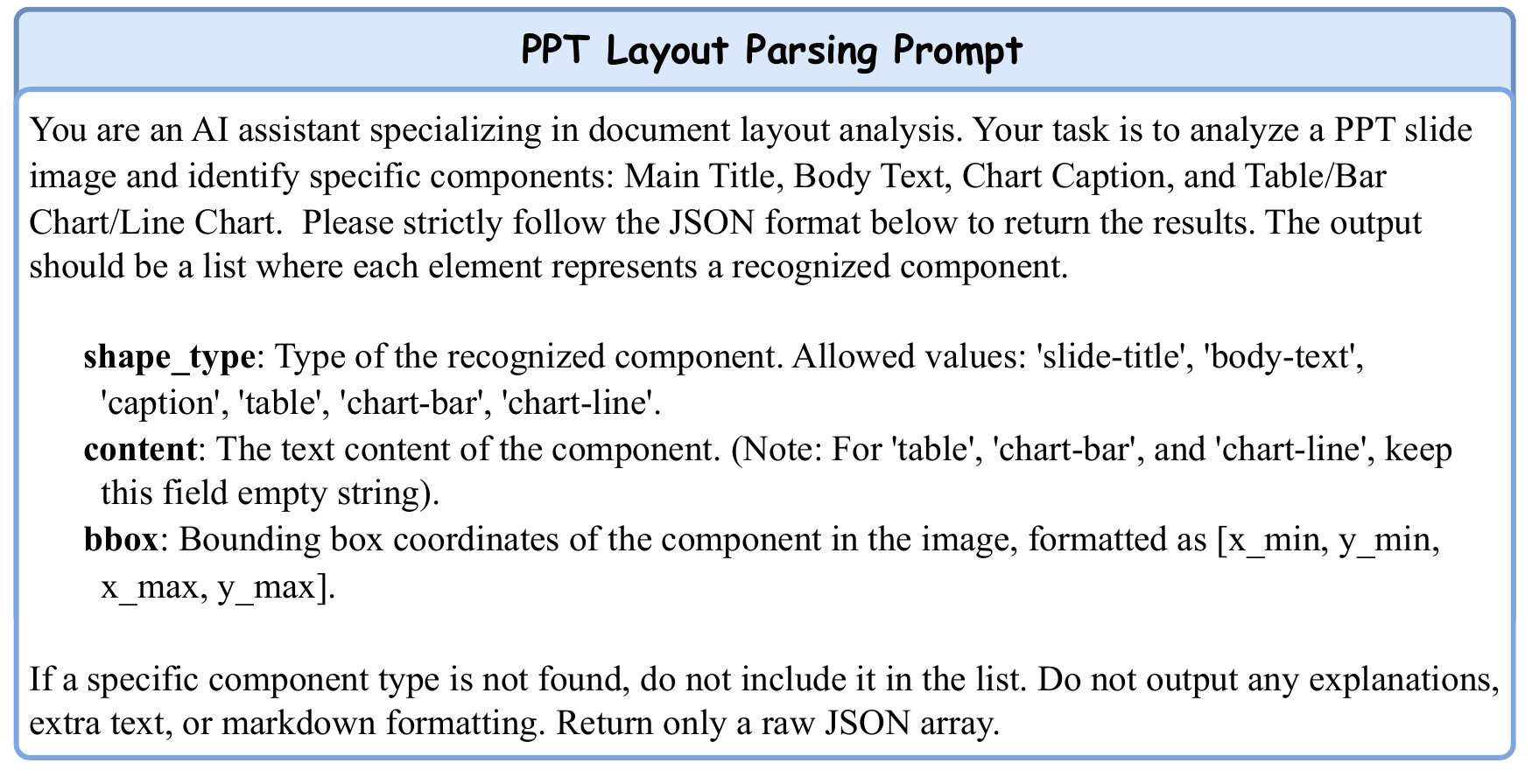}
  \caption{The prompt used for multimodal layout parsing to predict semantic labels and bounding boxes.}
  \label{fig:prompt_layout_parsing}
\end{figure}

\begin{figure}[t]
  \includegraphics[width=\columnwidth]{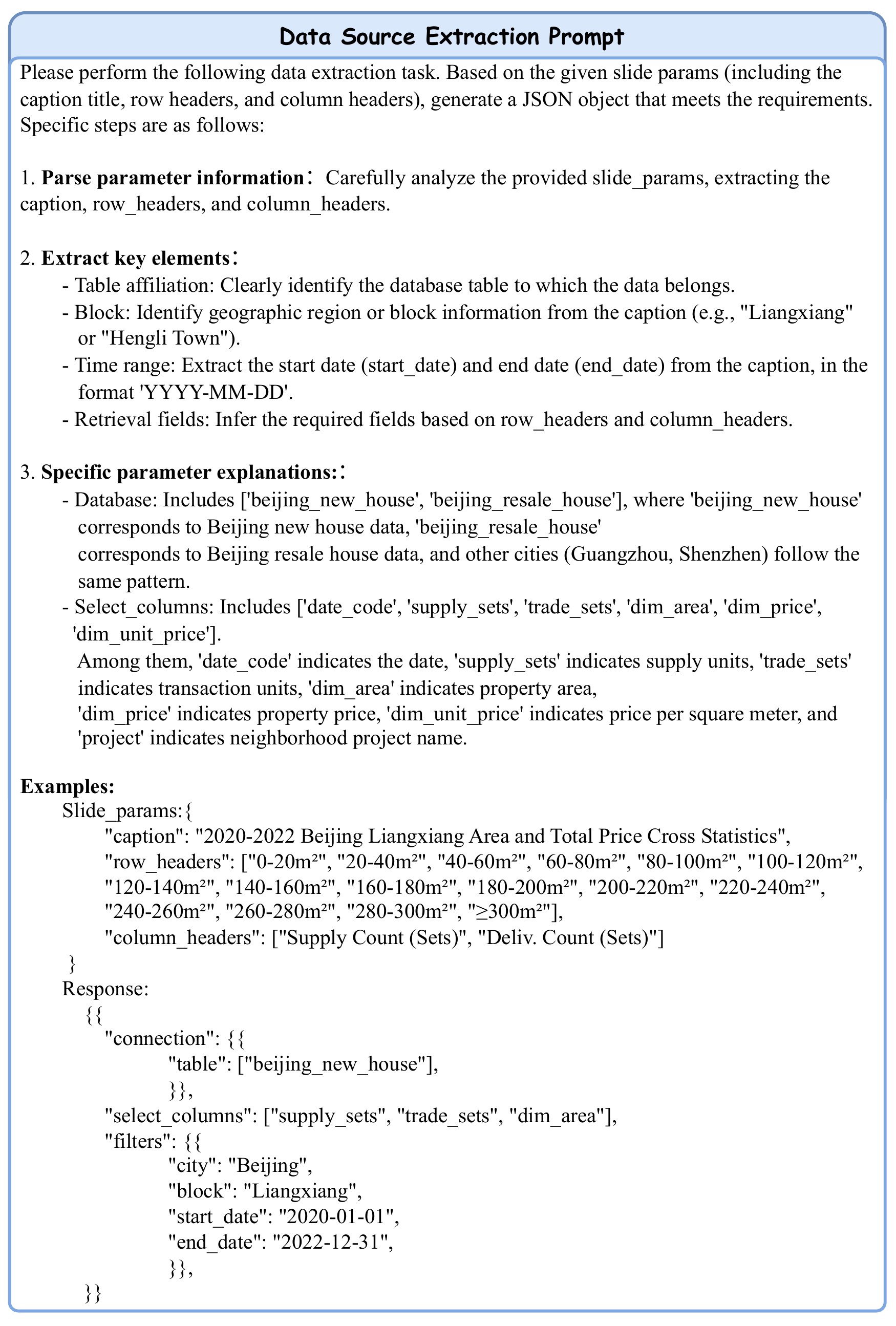}
  \caption{The prompt for data source extraction, mapping slide content to database schema slots.}
  \label{fig:prompt_data_source}
\end{figure}

\begin{figure}[t]
  \includegraphics[width=\columnwidth,height=1.0\textheight]{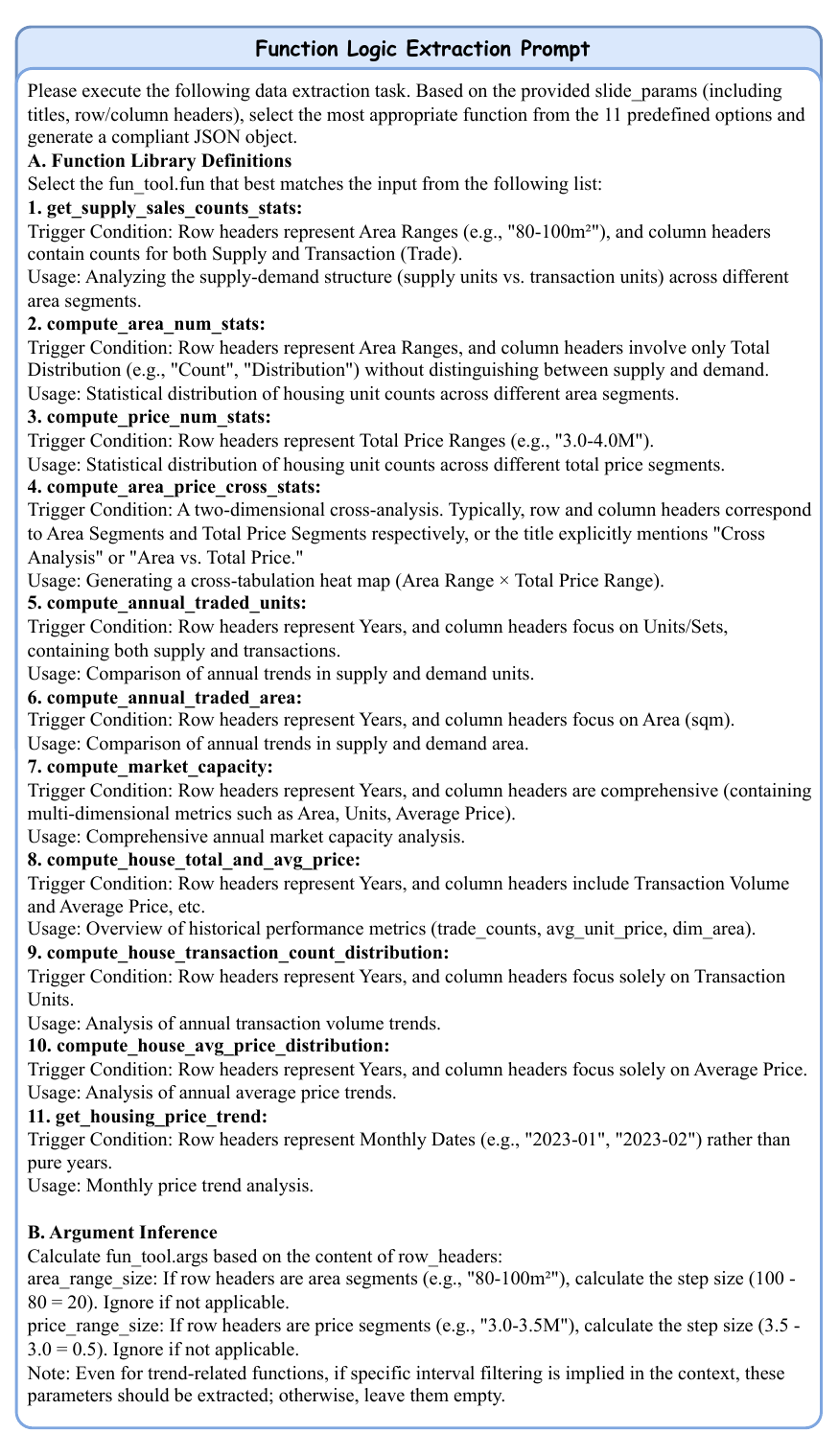}
  \caption{The prompt for closed-domain function logic extraction via predefined tool invocation.}
  \label{fig:prompt_function_logic_closed}
\end{figure}

\begin{figure}[t]
  \includegraphics[width=\columnwidth]{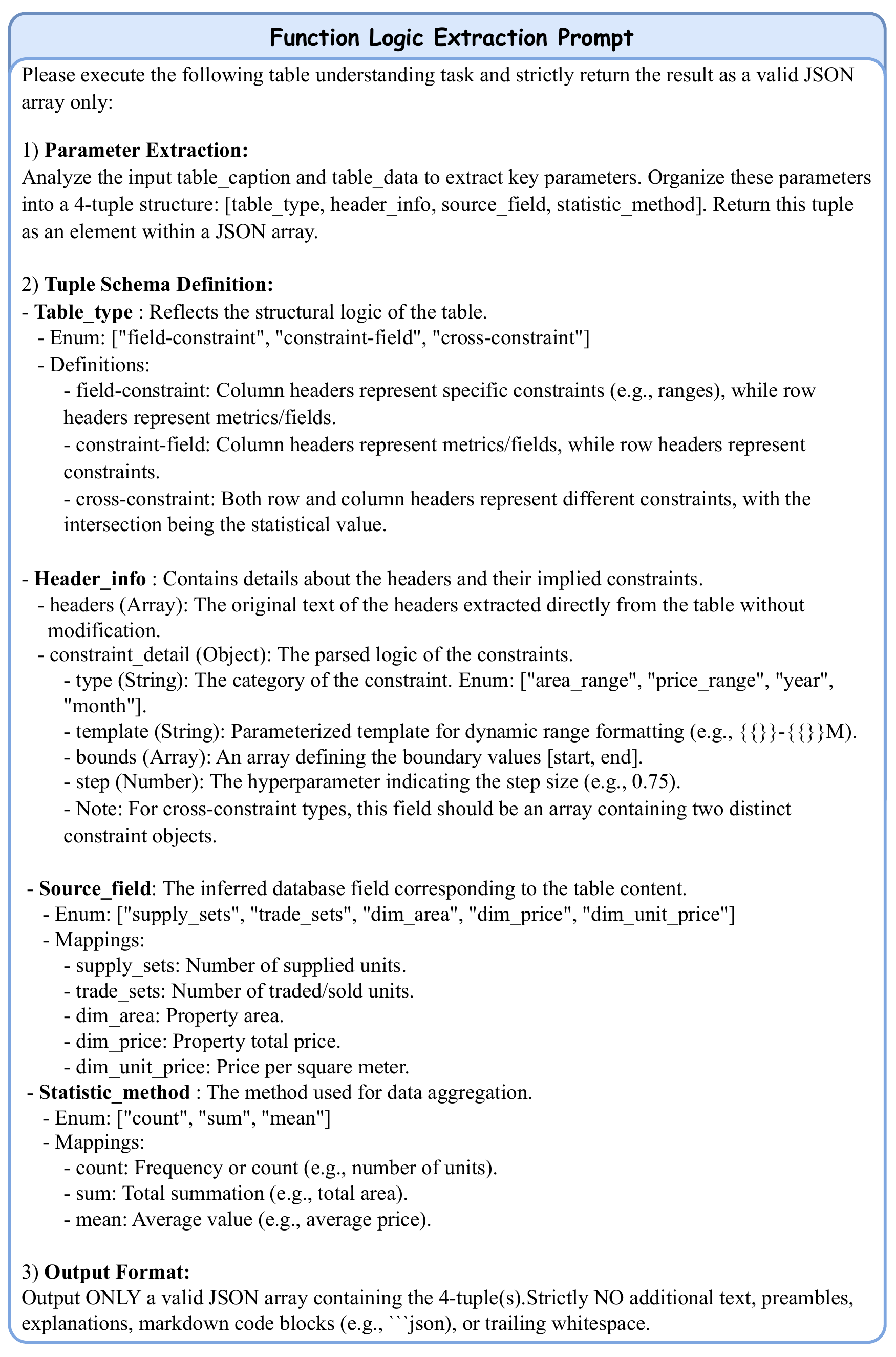}
  \caption{The prompt for open-domain logic extraction, decomposing unknown logic into atomic parameters.}
  \label{fig:prompt_function_logic_open}
\end{figure}

\begin{figure}[t]
  \includegraphics[width=\columnwidth]{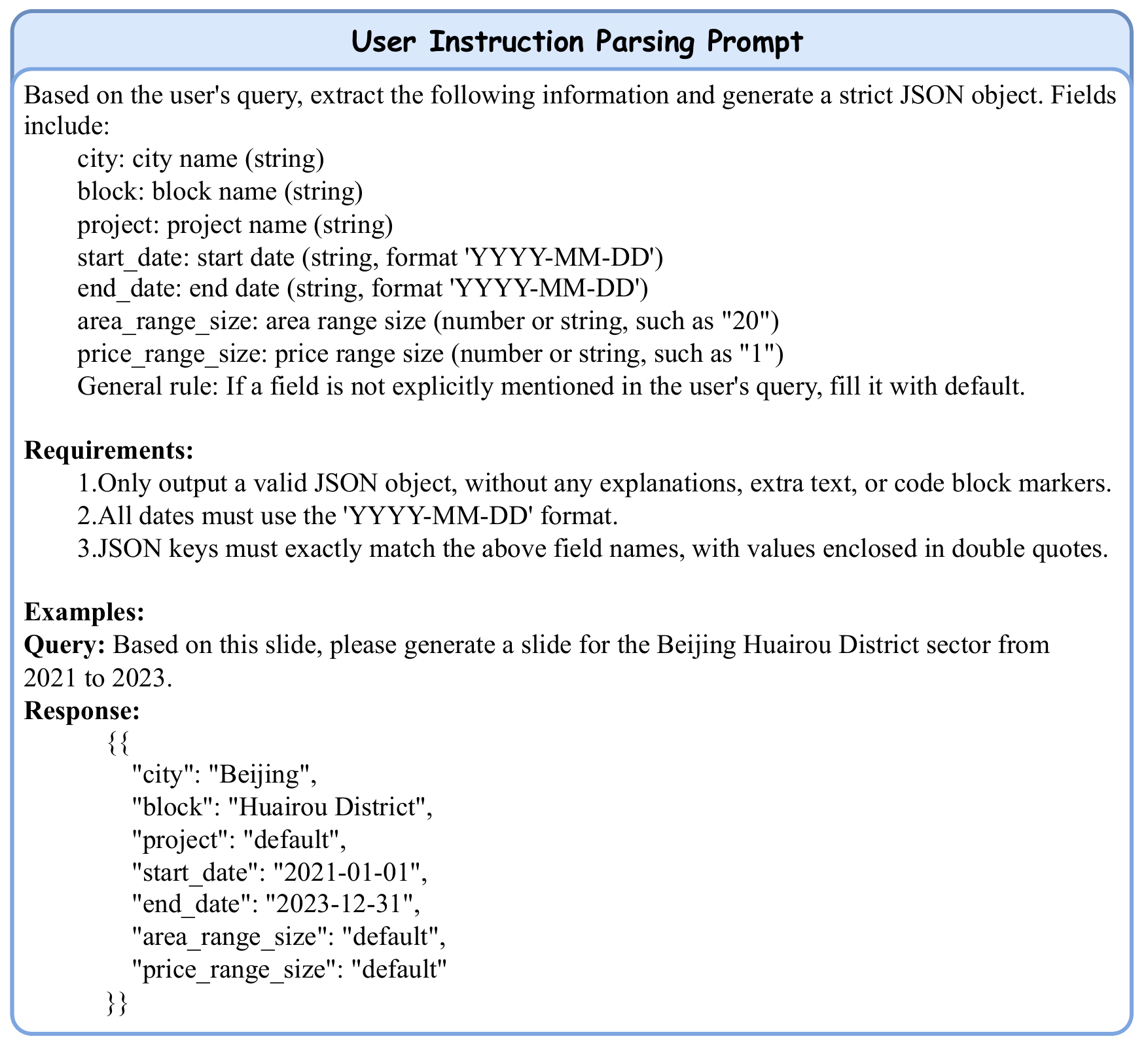}
  \caption{The prompt for user instruction parsing, modeled as a parameter state update task.}
  \label{fig:prompt_instruction_parsing}
\end{figure}

\begin{figure}[t]
  \includegraphics[width=\columnwidth]{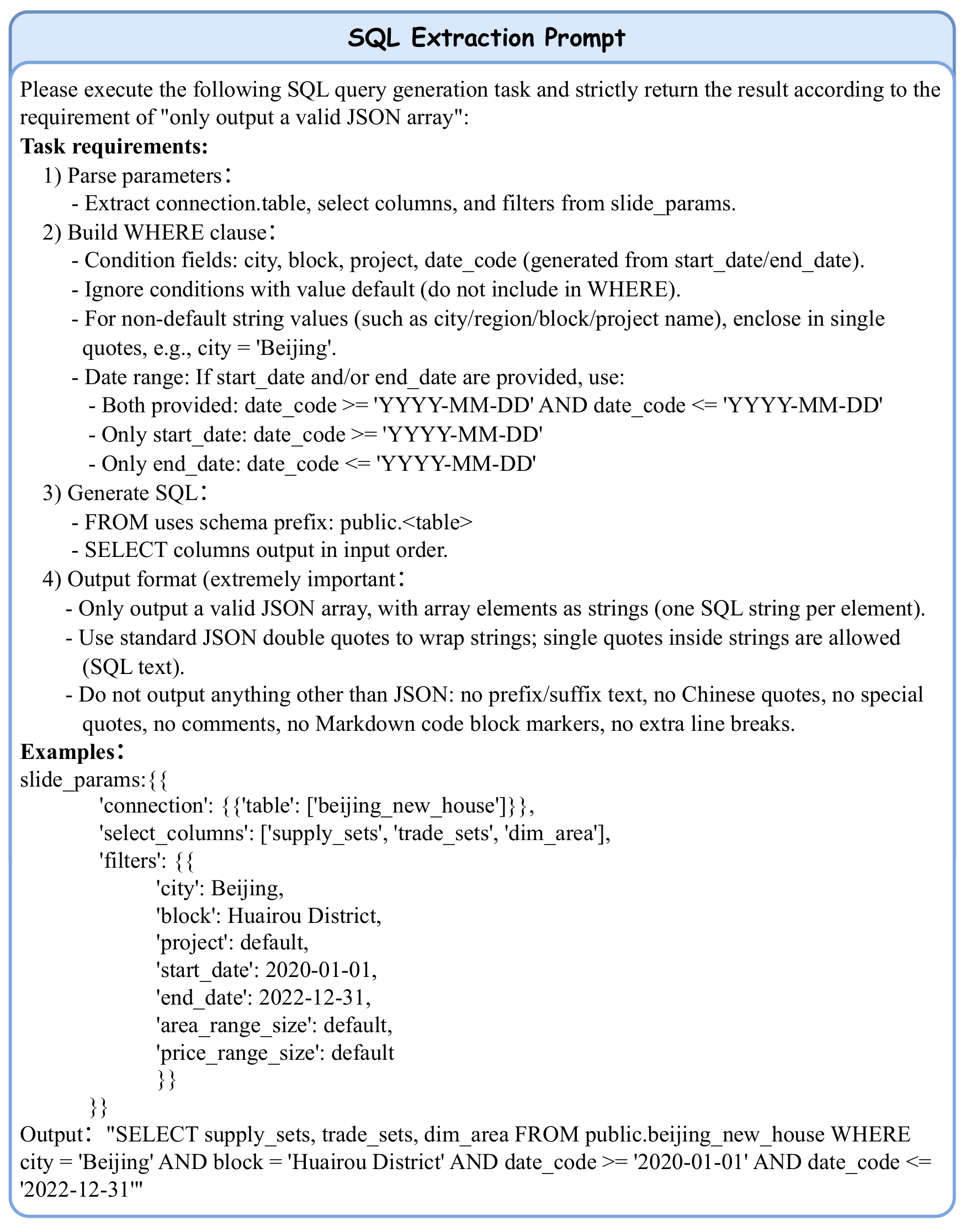}
  \caption{The prompt for SQL generation, converting updated parameters into executable database queries.}
  \label{fig:prompt_sql_generation}
\end{figure}


\begin{figure}[t]
  \includegraphics[width=\columnwidth,height=1.0\textheight]{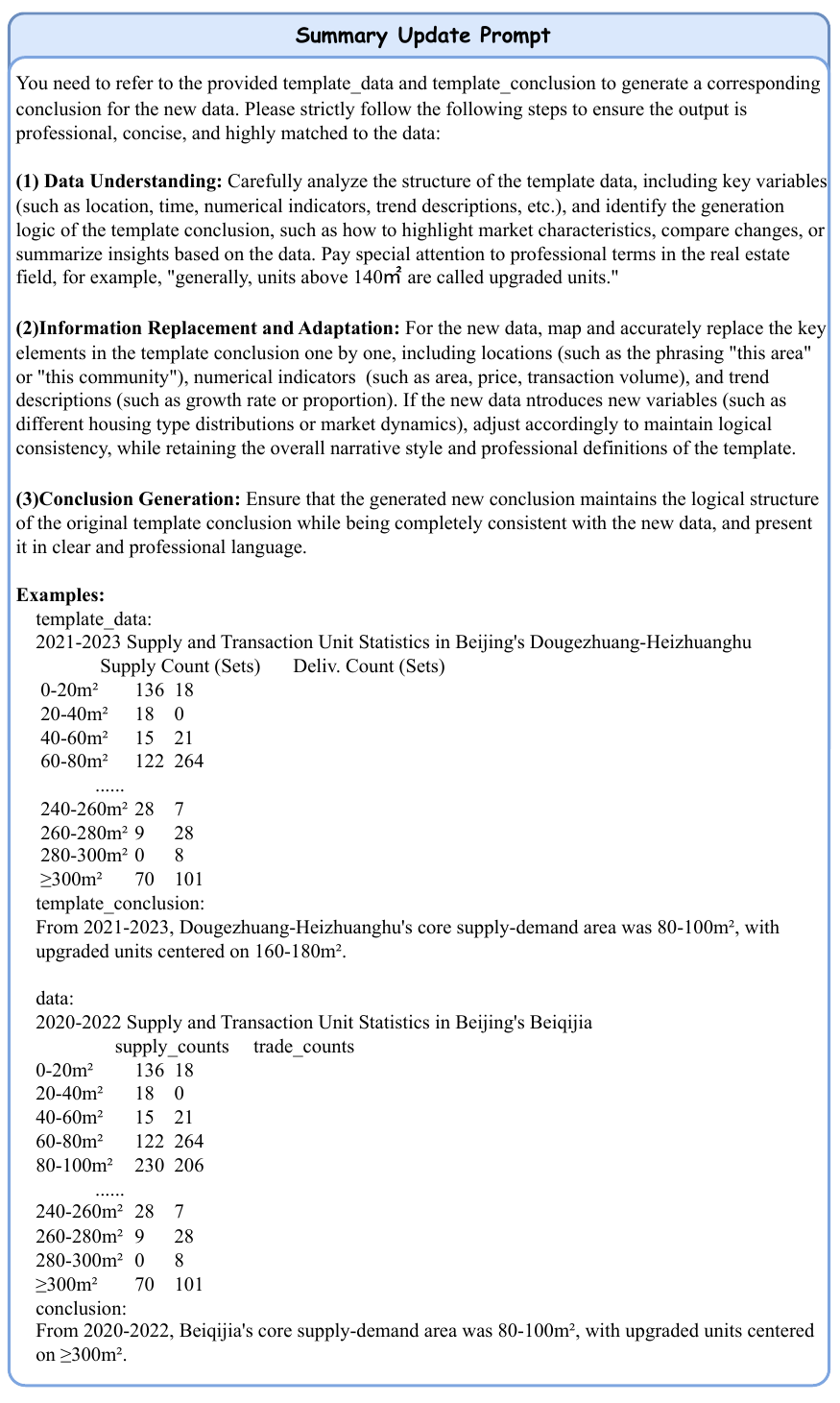}
  \caption{The prompt for fact-aware summary update based on data changes.}
  \label{fig:prompt_summary_update}
\end{figure}

\end{document}